\documentclass[sigconf,authorversion,balance=false]{acmart}

\AtBeginDocument{%
  }

\setcopyright{acmlicensed}
\copyrightyear{2026}
\acmYear{2026}
\setcopyright{acmlicensed}
\setcopyright{cc}
\setcctype{by}
\acmConference[KDD '26]{Proceedings of the 32nd ACM SIGKDD Conference on Knowledge Discovery and Data Mining V.2}{August 09--13, 2026}{Jeju Island, Republic of Korea}
\acmBooktitle{Proceedings of the 32nd ACM SIGKDD Conference on Knowledge Discovery and Data Mining V.2 (KDD '26), August 09--13, 2026, Jeju Island, Republic of Korea}
\acmDOI{10.1145/3770855.3817758}
\acmISBN{979-8-4007-2259-2/2026/08}

\usepackage[capitalize,noabbrev]{cleveref}
\usepackage{xspace}
\usepackage{mathtools}
\usepackage{bm}
\usepackage{subcaption}
\usepackage{stfloats}

\usepackage{algorithm}
\usepackage{algpseudocode}
\algrenewcommand\algorithmicrequire{\textbf{Input:}}
\algrenewcommand\algorithmicensure{\textbf{Output:}}

\makeatletter
\DeclareRobustCommand\onedot{\futurelet\@let@token\@onedot}
\def\@onedot{\ifx\@let@token.\else.\null\fi\xspace}
\makeatother
\newcommand\eg{e.g\onedot}
\newcommand\ie{i.e\onedot}
\newcommand\cf{cf\onedot}
\newcommand\vs{vs\onedot}
\newcommand\as{a.s\onedot}

\makeatletter
\newcommand{\AnonOnly}[1]{%
  \if@ACM@anonymous
    #1%
  \fi
}
\newcommand{\NonAnonOnly}[1]{%
  \if@ACM@anonymous
  \else
    #1%
  \fi
}
\newcommand{\AnonIfElse}[2]{%
  \if@ACM@anonymous
    #1%
  \else
    #2%
  \fi
}
\makeatother

\newcommand{\methodname}{%
  condPED-ANOVA\xspace%
}

\newcommand{\CondPedAnovaGitHubLink}{%
  \AnonIfElse{%
    \url{https://anonymous.4open.science/r/condPED-ANOVA-anonymous}%
  }{%
    \url{https://github.com/kAIto47802/condPED-ANOVA}%
  }%
  \xspace
}

\makeatletter
\patchcmd{\maketitle}
  {\@specialsection{\keywordsname}}
  {\vspace{-7pt}\@specialsection{\keywordsname}}
  {}{}
\makeatother

\begin{document}

\title[Conditional PED-ANOVA: Hyperparameter Importance in Hierarchical \& Dynamic Search Spaces]{
  Conditional PED-ANOVA: Hyperparameter Importance\\in Hierarchical \& Dynamic Search Spaces
}

\author{Kaito Baba}
\authornote{This work was conducted at Preferred Networks, Inc.}
\orcid{0009-0004-1351-6097}
\affiliation{%
  \institution{Preferred Networks, Inc.\\The University of Tokyo}
  \city{Tokyo}
  \country{Japan}
}
\email{baba-kaito662@g.ecc.u-tokyo.ac.jp}

\author{Yoshihiko Ozaki}
\orcid{0000-0001-7911-4470}
\affiliation{%
  \institution{Preferred Networks, Inc.}
  \city{Tokyo}
  \country{Japan}
}
\email{yozaki@preferred.jp}

\author{Shuhei Watanabe}
\authornote{This work was conducted while the author was affiliated with Preferred Networks, Inc.}
\orcid{0009-0001-4252-6612}
\affiliation{%
  \institution{SB Intuitions Corp.}
  \city{Tokyo}
  \country{Japan}
}
\email{shuhei.watanabe@sbintuitions.co.jp}

\renewcommand{\shortauthors}{Kaito Baba, Yoshihiko Ozaki, \& Shuhei Watanabe}

\begin{abstract}
We propose \emph{conditional PED-ANOVA (\methodname)}, a principled framework for estimating hyperparameter importance (HPI) in conditional search spaces, where the presence or domain of a hyperparameter can depend on other hyperparameters.
Although the original PED-ANOVA provides a fast and efficient way to estimate HPI within the top-performing regions of the search space, it assumes a fixed, unconditional search space and therefore cannot properly handle conditional hyperparameters.
To address this, we introduce a conditional HPI for top-performing regions and derive a closed-form estimator that accurately reflects conditional activation and domain changes.
Experiments show that naive adaptations of existing HPI estimators yield misleading or uninterpretable importances in conditional settings, whereas \methodname consistently provides meaningful importances that reflect the underlying conditional structure.
Our code is publicly available at \CondPedAnovaGitHubLink.
\end{abstract}




\maketitle

\section{Introduction}
\label{sec:intro}

Analysis of parameter importance~\citep{sobol1993sensitivity,sobol2001global,hooker2007generalized,huang1998projection} is a central tool for understanding and designing optimization procedures.
Over the past decade, techniques for measuring parameter importance have attracted considerable attention in the automated machine learning (AutoML) community~\citep{hutter2014efficient,Biedenkapp_Lindauer_Eggensperger_Hutter_Fawcett_Hoos_2017,van2018hyperparameter}, particularly for improving hyperparameter optimization (HPO)~\citep{NIPS2011_86e8f7ab,JMLR:v13:bergstra12a,NIPS2012_05311655,JMLR:v23:21-0888,eggensperger2013towards}.
Given a hyperparameter search space and evaluated configurations, methods for hyperparameter importance (HPI) aim to quantify how strongly each hyperparameter influences the observed performance.
Such information helps interpret optimization runs, identify critical design choices, detect irrelevant or redundant parameters, and guide subsequent search or manual tuning~\citep{melis2018on,10.5555/3504035.3504427}.
As a result, HPI is widely used across various settings, including HPO of machine learning models and HPO algorithm design, where understanding hyperparameter sensitivity is often essential~\citep{van2018hyperparameter,9382913,JMLR:v20:18-444,Theodorakopoulos_2024}.

Classic HPI methods such as functional analysis of variance (f-ANOVA)~\citep{hutter2014efficient} quantify HPI by measuring how much objective variance is attributed to each hyperparameter.
However, their existing local variants mix contributions from unpromising regions, leading to inaccurate estimates of HPI in top-performing subspaces~\citep{hutter2014efficient,10.1007/978-3-030-05348-2_10}.
PED-ANOVA~\citep{watanabe2023ped} addresses this by defining local HPI within a top-performing subset and efficiently approximating it using the Pearson divergence between one-dimensional distributions of each hyperparameter.
Moreover, PED-ANOVA computes both global and local HPIs substantially faster than f-ANOVA while preserving accuracy.
Thanks to this combination of speed and reliability, it has been adopted for HPI analysis in practical HPO frameworks, such as Optuna~\citep{akiba2019optuna}, demonstrating its utility in real-world workflows.

However, we argue that PED-ANOVA has a practical limitation: it cannot properly handle \emph{hierarchical}/\emph{dynamic} search spaces (hereafter \emph{conditional} search spaces), where the presence or domain of a hyperparameter depends on others.
In practice, many real-world problems involve conditional search spaces~\citep{10.5555/3360092}.
Typical examples include model selection, where hyperparameters depend on the chosen algorithm~\citep{10.1145/2487575.2487629,JMLR:v18:16-261,NIPS2015_11d0e628,komer-proc-scipy-2014}, and neural network architecture design, where active parameters depend on the chosen layers~\citep{bergstra2013making,swersky2014raiderslostarchitecturekernels}.
Unfortunately, PED-ANOVA cannot be directly applied to such settings, as it assumes a fixed, unconditional search space and defines distributions over a single, fixed domain for each hyperparameter.
We observe that naive extensions of PED-ANOVA, such as filtering inactive samples, imputing missing values, or artificially expanding parameter domains, do not resolve this issue: these heuristics often violate the conditional semantics of the search space and yield unnatural or uninterpretable HPI estimates (\cf \cref{sec:comparison-with-existing-methods}).

To address this limitation, we propose \emph{conditional PED-ANOVA (\methodname)}, a principled framework for estimating HPI in conditional search spaces.
We identify a critical flaw in PED-ANOVA's original definition of local HPI under conditionality: the importance of hyperparameters that induce conditionality is spuriously attributed to dependent hyperparameters, yielding systematically misleading HPI estimates (empirically shown in \cref{sec:ablation-within-regime-variance} and theoretically in \cref{thm:gating-leakage}).
Consequently, merely constructing a method that correctly computes local HPI in conditional search spaces is still insufficient.
We thus introduce \emph{conditional local HPI}, which measures importance within top-performing regions without confounding from conditional activation or domain changes.
Building on this definition, we derive a closed-form PED-ANOVA-style estimator that yields interpretable HPIs in conditional settings.
Experiments confirm that \methodname reliably produces a meaningful HPI that reflects the underlying conditional structure, while naive extensions of existing methods fail to do so.

Our contributions are summarized as follows:
\begin{itemize}
  \item We identify a fundamental pitfall of original local HPI in conditional search spaces: the contribution of conditioning hyperparameters is spuriously attributed to conditionally active (or domain-shifted) hyperparameters, yielding misleading importance estimates.
  \item We introduce conditional local HPI, a definition that measures importance without being confounded by conditional activation and domain changes.
  \item \sloppypar{We derive a closed-form PED-ANOVA-style estimator, namely \methodname, for conditional local HPI, retaining PED-ANOVA's efficiency.}
  \item We empirically demonstrate that \methodname consistently produces meaningful importances in conditional settings, whereas naive extensions of existing methods, including PED-ANOVA, produce unnatural results.
\end{itemize}

\section{Related Work}
\label{sec:related-work}

Here, we review prior work most relevant to ours.
Additional related work is provided in \cref{apx:additional_related_work}.

\subsection{Variance-Based HPI and f-ANOVA}

HPI aims to quantify how strongly each hyperparameter (and their interactions) influences performance, typically to support search-space design, debugging, and interpretation of HPO runs.
Early work on parameter importance analysis builds on variance-based sensitivity analysis such as Sobol indices~\citep{sobol1993sensitivity,sobol2001global} and ANOVA decompositions~\citep{hooker2007generalized,huang1998projection}, which attribute the global variance of an objective to individual coordinates and their interactions.
In the context of HPO, f-ANOVA~\citep{hutter2014efficient} applies this idea to HPO objectives by fitting a surrogate model (\eg, a random forest) over the search space and estimating HPIs by integrating the surrogate over tree partitions using region-volume weights.
This produces global importance scores that quantify how much of the performance variance is explained by each hyperparameter, and it has become a standard post-hoc analysis tool in AutoML.

\subsection{PED-ANOVA}

PED-ANOVA~\citep{watanabe2023ped} extends HPI to settings where the region of interest is not an entire search space.
It defines local HPI as the variance contribution of each hyperparameter within the top-$\gamma$ subset of the search space, thereby measuring importance directly in the most relevant region.
It further shows that this local HPI can be approximated in closed form using the Pearson divergence between the one-dimensional distributions of each hyperparameter of the two top-performing subsets, estimated via one-dimensional kernel density estimators (KDEs)~\citep{parzen1962estimation}.
Empirically, this leads to much faster HPI computation than f-ANOVA while maintaining accurate importance estimates in top-performing subspaces.
The details of PED-ANOVA are reviewed in \cref{sec:preliminaries}.

Our approach follows a PED-ANOVA-style framework, while enabling effective HPI estimation in conditional search spaces that arise in many real-world problem settings.

\subsection{Conditional Hyperparameters in HPO}

Many HPO algorithms operate on conditional search spaces, where some hyperparameters' activation and domains depend on specific parent choices (\eg, algorithm selection).
This conditional structure is explicitly modeled in practical AutoML systems, such as Auto-WEKA~\citep{10.1145/2487575.2487629} and auto-sklearn~\citep{NIPS2015_11d0e628,10.5555/3586589.3586850}.
Among widely used optimizers, SMAC~\citep{JMLR:v23:21-0888} is designed to handle such conditional hierarchies in CASH-style spaces~\citep{10.1145/2487575.2487629}, while TPE~\citep{NIPS2011_86e8f7ab} formulates the search space as a tree-structured generative process.
BOHB~\citep{pmlr-v80-falkner18a} combines Hyperband~\citep{JMLR:v18:16-558} with a TPE-based Bayesian optimization (BO), optimizing over the same tree-structured conditional spaces while exploiting cheap low-fidelity evaluations.
Beyond these algorithms, a complementary work develops BO surrogates tailored to conditional domains to share information across branches~\citep{swersky2014raiderslostarchitecturekernels,pmlr-v70-jenatton17a,pmlr-v108-ma20a}.

\subsection{HPI under Conditional Search Spaces}

In conditional search spaces, applying standard HPI pipelines requires mapping configurations to a fixed representation, often by imputing inactive hyperparameters with default values as done in SMAC-style toolchains~\citep{JMLR:v23:21-0888}.
Practical analysis suites such as CAVE~\citep{10.1007/978-3-030-05348-2_10} (and its integration in BOAH~\citep{lindauer2019boahtoolsuitemultifidelity}) and DeepCAVE~\citep{c41e4530763a48718e9783a19c72df78} then compute importance post-hoc using methods like f-ANOVA and local parameter importance, including for multi-fidelity runs.
Current Optuna's importance API highlights a common limitation of conditional spaces: by default, it evaluates only parameters that appear and have the same domain in all samples, excluding conditional parameters~\citep{akiba2019optuna}.
Recent work on multi-objective HPI likewise treats missing values arising from conditional hyperparameters in preprocessing before applying surrogate-based importance measures such as f-ANOVA and ablation paths~\citep{Theodorakopoulos_2024}.

Consequently, despite the fact that conditional hyperparameters arise frequently in practical HPO settings, prior HPI methods have lacked a principled way to handle conditional hyperparameters directly, without resorting to default-value imputation or discarding them during preprocessing. This limitation has prevented existing approaches from faithfully capturing the importance of hyperparameters within the conditional structure of the search space.

\section{Preliminaries}
\label{sec:preliminaries}

\subsection{Notations}

Let $\smash{\mathcal{X} \coloneqq \mathcal{X}^{(1)} \times \cdots \times \mathcal{X}^{(D)}}$ be the $D$-dimensional search space for hyperparameters $\smash{\bm x \coloneqq (x^{(1)}, \ldots, x^{(D)})\in\mathcal{X}}$.
Let $f\colon\mathcal X\to\mathbb R$ be an objective function to be minimized.
We observe a finite evaluation set $\smash{\{(\bm{x}_n,f(\bm{x}_n))\}_{n=1}^N}$, where each $\bm x_n$ is sampled from a distribution $p_0(\bm x)$.
For $\gamma\in(0,1]$, let $f_\gamma\in\mathbb{R}$ denote the lower (\ie, better) empirical $\gamma$-quantile of the
observed values, 
\ie, $\smash{f_\gamma}$ satisfies
$
  \#\{n \mid f(\bm{x}_n)\leq f_\gamma\} = \lfloor\gamma N\rfloor.
$
The corresponding top-$\gamma$ set is represented as
$
  \mathcal{X}_\gamma := \{\bm x\in\mathcal X \mid f(\bm x)\leq f_\gamma\}.
$

\subsection{Local HPI}

To quantify how strongly the $d$-th hyperparameter influences the performance in the top-$\gamma$ set, a local HPI based on a local marginal variance is introduced~\citep{watanabe2023ped}.
Let
$
  b_\gamma(\bm x) \coloneqq \mathbf 1\{\bm x\in \mathcal X_\gamma\} = \mathbf 1\{f(\bm x)\leq f_\gamma\}
$
be the indicator function of whether hyperparameters $\bm x$ belong to the top-$\gamma$ performers.
The \textit{local mean} $m_\gamma$ of the objective and the \textit{local marginal mean} $\smash{f_\gamma^{(d)}(x^{(d)})}$ for the $d$-th hyperparameter $\smash{x^{(d)}}$ at level $\gamma$ are defined as:
\begin{align}
  &\hspace{9.53mm}
  m_\gamma
  \coloneqq
  \mathbb{E}_{\bm x \sim p_0}[f(\bm x)\mid b_\gamma(\bm x)]
  =
  \frac{\mathbb{E}_{\bm x \sim p_0}[f(\bm x)\, b_\gamma(\bm x)]}{\mathbb{E}_{\bm x \sim p_0}[b_\gamma(\bm x)]},
  \\
  &\begin{aligned}
    f_\gamma^{(d)}(x^{(d)})
    &\coloneqq
    \mathbb{E}_{\bm x^{(\bar d\,)} \sim p_0^{(\bar d\,)}}
    [f(\bm x)\mid b_\gamma(\bm x), x^{(d)}]
    \\
    &\,=
    \frac{
      \mathbb{E}_{\bm x^{(\bar d\,)} \sim p_0^{(\bar d\,)}}
      [f(\bm x)\, b_\gamma(\bm x)\mid x^{(d)}]
    }{
      \mathbb{E}_{\bm x^{(\bar d\,)} \sim p_0^{(\bar d\,)}}
      [b_\gamma(\bm x)\mid x^{(d)}]
    },
  \end{aligned}
\end{align}
where
$\smash{\bm x^{(\bar d\,)}} \!\coloneqq\! (\smash{x^{(1)}},\ldots,\smash{x^{(d-1)}},\smash{x^{(d+1)}},\ldots,\smash{x^{(D)}})$
denotes all hyperparameters except for the $d$-th one.
The local marginal mean $f_\gamma^{(d)}(x^{(d)})$ is obtained by taking the mean over all hyperparameters $\smash{\bm x^{(\bar d\,)}}$ within the top-$\gamma$ region while keeping the $d$-th hyperparameter fixed at $\smash{x^{(d)}}$, in a manner similar to marginalization.

The \textit{local marginal variance} for the $d$-th hyperparameter at level $\gamma$ is then defined as follows:
\begin{equation}
  \begin{aligned}
    \smash{v_\gamma^{(d)}}
    &\coloneqq
    \mathrm{Var}_{\bm x \sim p_0}
    \bigl(\smash{f_\gamma^{(d)}(x^{(d)})} \mid b_\gamma(\bm x)\bigr)
    \\
    &\,=
    \frac{
      \mathbb{E}_{\bm x \sim p_0}[
        (f_\gamma^{(d)}(x^{(d)}) - m_\gamma)^2\, b_\gamma(\bm x)
      ]
    }{
      \mathbb{E}_{\bm x \sim p_0}
      [b_\gamma(\bm x)]
    }.
  \end{aligned}
\end{equation}
Using this quantity, the \textit{local HPI} for the $d$-th hyperparameter at level $\gamma$ is then defined as
its normalized form:
\begin{equation}
  \text{Local HPI} \coloneqq v_\gamma^{(d)}\Big/\sum_{d'=1}^D v_\gamma^{(d')}.
\end{equation}

\subsection{PED-ANOVA}

PED-ANOVA~\citep{watanabe2023ped} introduces an efficient approximation of the local HPI of the objective $f$ by
computing the local HPI of a tighter level-set indicator function
$
  b_{\gamma^\prime}(\bm x) \coloneqq \mathbf{1}\{\bm x\in\mathcal{X}_{\gamma^\prime}\},
$
where $0 < \gamma^\prime < \gamma$.
A key result in PED-ANOVA shows that the local marginal variance for $b_{\gamma^\prime}$
admits the following closed-form expression:
\begin{equation}
  v_\gamma^{(d)}
  =
  \left(\frac{\gamma^\prime}{\gamma}\right)^{\!2}
  D_{\mathrm{PE}}\left(
    p_{\gamma^\prime}^{(d)} \,\middle\|\, p_\gamma^{(d)}
  \right),
  \label{eq:ped-local-hpi}
\end{equation}
where $\smash{p_{\gamma}^{(d)}}$ and $\smash{p_{\gamma^\prime}^{(d)}}$ are the empirical
one-dimensional probability density functions (PDFs) of the $d$-th hyperparameter induced by the samples in the top-$\gamma$ and top-$\gamma^\prime$ subsets, respectively.
In practice, the PDFs are estimated via KDE using the observed samples.
Here, $D_{\mathrm{PE}}(p\|q)$ denotes the Pearson ($\chi^2$) divergence between two PDFs $p$ and $q$, defined as:
\begin{equation}
  \smash{D_{\mathrm{PE}}(p\|q)
  \coloneqq
  \int
  \left(
    \frac{p(x)}{q(x)} - 1
  \right)^2
  q(x)
  \mathrm{d}x.}
\end{equation}
Note that this divergence is well-defined in the present setting, since $\smash{\mathcal{X}_{\gamma^\prime} \subset \mathcal{X}_\gamma}$ implies that the density $\smash{p_{\gamma^\prime}^{(d)}}$ has support only where $p_{\gamma}^{(d)}$ is nonzero, so the ratio $\smash{p_{\gamma^\prime}^{(d)} / p_{\gamma}^{(d)}}$ is well-defined.

PED-ANOVA uses this closed-form expression to compute the local HPI solely from one-dimensional PDFs.
In the original PED-ANOVA experiments, this approach was shown to yield dramatically faster HPI calculation compared to f-ANOVA, while maintaining comparable quality in the resulting importance measures.

\section{\methodname: Hyperparameter Importance in Conditional Search Spaces}

\label{sec:c-ped-anova}

In this section, we present \methodname, a principled approach for efficient HPI computation in conditional search spaces.
Rigorous details are provided in \cref{apx:c-ped-anova-details}.

\subsection{Regime-Based Representation of Conditional Hyperparameters}
\label{sec:regime-based-representation}

In practice, many hyperparameters are conditional: the presence
of a hyperparameter and its domain can change depending on other hyperparameters~\citep{10.5555/3360092,10.1145/2487575.2487629,JMLR:v18:16-261,NIPS2015_11d0e628,komer-proc-scipy-2014,bergstra2013making,swersky2014raiderslostarchitecturekernels}.

We now formalize such behavior.
To handle such conditional structure, we partition the behavior of the $d$-th hyperparameter into a finite number of \emph{regimes}.
Each regime $\smash{i^{(d)}\in\{1,\ldots,K^{(d)}\}}$ has its own domain $\smash{\mathcal{Z}_i^{(d)}}$ for $\smash{x^{(d)}}$, which may differ across regimes.
When $\smash{x^{(d)}}$ is inactive in regime $\smash{i^{(d)}}$, we represent this by a special symbol $\bot$ and set $\smash{\mathcal{Z}_i^{(d)}=\{\bot\}}$; otherwise, $\smash{\mathcal{Z}_i^{(d)}}$ is the usual domain (e.g., an interval or a discrete set), whose range may depend on $\smash{i^{(d)}}$.

For example, suppose $\smash{x^{(0)}}$ selects a learning algorithm, such as a neural network or a tree-based model.
If $\smash{x^{(0)}=\texttt{neural}}$, $\smash{x^{(1)}}$ may correspond to the number of layers with domain $\smash{\mathcal{Z}^{(1)}_{\texttt{neural}}}=\{1,\ldots,L_{\max}\}$, while $\smash{x^{(2)}}$ is inactive, \ie, $\smash{\mathcal{Z}^{(2)}_{\texttt{neural}}=\{\bot\}}$.
Conversely, if $x^{(0)}=\texttt{tree}$, $x^{(1)}$ is inactive, \ie, $\smash{\mathcal{Z}^{(1)}_{\texttt{tree}}=\{\bot\}}$, and $\smash{x^{(2)}}$ may represent the minimum split gain with domain $\smash{\mathcal{Z}^{(2)}_{\texttt{tree}}=\mathbb{R}_{+}}$.
Here, the choice of $\smash{x^{(0)}}$ induces regimes $\smash{i^{(d)}\in\{\texttt{neural},\texttt{tree}\}}$ under which $\smash{x^{(1)}}$ and $\smash{x^{(2)}}$ have regime-specific domains or become inactive.

\begin{algorithm*}[t]
  \caption{\methodname Computation for the $d$-th Hyperparameter $x^{(d)}$}
  \label{alg:cond-ped-anova}
  \begin{algorithmic}[1]
    \Require Evaluation set $\mathcal D = \{(\bm x_n, f(\bm x_n))\}_{n=1}^N$, hyperparameter index $d$, quantile levels $0 < \gamma^\prime \le \gamma \le 1$
    \Ensure The within-regime local marginal variance $v_{\gamma,\mathrm{within}}^{(d)}$ for the $d$-th hyperparameter $x^{(d)}$
    \State Compute empirical $\gamma^\prime$- and $\gamma$- quantiles $f_{\gamma^\prime}$ and $f_{\gamma}$ of $\{f(\bm x_n)\}_{n=1}^N$
    \State Define top-performing subsets:
           $\mathcal D_{\gamma^\prime} \gets \{(\bm x_n,f(\bm x_n))\mid f(\bm x_n) \le f_{\gamma^\prime}\}$,
           $\mathcal D_{\gamma} \gets \{(\bm x_n,f(\bm x_n))\mid f(\bm x_n) \le f_{\gamma}\}$
    \For{$i = 1,\ldots,K^{(d)}$} \Comment{process each regime $i$}
      \State $\mathcal{D}_{\gamma^\prime,\,i}^{(d)}\,,\, \mathcal{D}_{\gamma,\,i}^{(d)} \gets$ Extract samples in regime $i$ from $\mathcal{D}_{\gamma^\prime}$ and $\mathcal{D}_{\gamma}$
      \State $p_{\gamma^\prime,\,i}^{(d)}\,,\, p_{\gamma,\,i}^{(d)} \gets$ Estimate via one-dimensional KDE from the samples $\mathcal{D}_{\gamma^\prime,\,i}^{(d)}$ and $\mathcal{D}_{\gamma,\,i}^{(d)}$
      \State $L^\text{within}_i \gets D_{\mathrm{PE}}\left(p_{\gamma^\prime,\,i}^{(d)} \,\middle\|\,p_{\gamma,\,i}^{(d)}\right)$\Comment{within-regime Pearson divergence}
      \State $
        \alpha_i^{(d)} \gets |\mathcal{D}_{\gamma^\prime,\,i}^{(d)}| \,/\, |\mathcal{D}_{\gamma^\prime}|
        ,\;
        \beta_i^{(d)} \gets |\mathcal{D}_{\gamma,\,i}^{(d)}| \,/\, |\mathcal{D}_{\gamma}|
      $
      \Comment{regime probabilities in top $\gamma^\prime$ and $\gamma$ subsets}
    \EndFor
    \State \Return $
      \left(\frac{\gamma^\prime}{\gamma}\right)^2
      \sum_{i=1}^{K^{(d)}} \frac{\big(\alpha_i^{(d)}\big)^2}{\beta_i^{(d)}}L^\text{within}_i
    $
    \Comment{weighted aggregation of within-regime divergences}
  \end{algorithmic}
  \vspace{-2pt}
\end{algorithm*}

\subsection{Conditional Local HPI via Within-Regime Variance}
\label{sec:within-regime-hpi}

Let $\smash{I^{(d)}}$ and $\smash{Z^{(d)}}$ denote random variables taking values in $\{1,\ldots,\allowbreak K^{(d)}\}$ and $\mathcal{Z}^{(d)}_{I^{(d)}}$, respectively.
We write the local marginal mean $\smash{f_\gamma^{(d)}(x^{(d)})}$ as
$
  g_\gamma^{(d)}(I^{(d)}, Z^{(d)}).
$

We observe that the standard local HPI based on the local marginal variance $v_{\gamma}^{(d)}$ mixes two distinct sources of variation:
differences \emph{within} each regime and differences \emph{between} (inter) regimes.
This can be seen through the following variance decomposition by the law of total variance:
\begin{equation}
  \begin{aligned}
    v_{\gamma}^{(d)}
    &=
    \mathrm{Var}_{I^{(d)},\,Z^{(d)}}\left(g_\gamma^{(d)}(I^{(d)},Z^{(d)})\right)
    \\
    &=
    \underbrace{
      \mathbb{E}_{I^{(d)}} \left[
        \mathrm{Var}_{Z^{(d)}} \left(g_\gamma^{(d)}(I^{(d)},Z^{(d)})\,\middle|\, I^{(d)}\right)
      \right]
    }_{
      \text{within-regime variance}
    }
    \\
    &\hspace{2.6em}+
    \underbrace{
      \mathrm{Var}_{I^{(d)}} \left(
        \mathbb{E}_{Z^{(d)}} \left[g_\gamma^{(d)}(I^{(d)},Z^{(d)})\,\middle|\, I^{(d)}\right]
      \right)
    }_{
      \text{inter-regime variance}
    },
  \end{aligned}
  \label{eq:var-decomp-within-inter}
\end{equation}
where the expectations and variances are taken under the empirical distribution restricted to the top-$\gamma$ region $\mathcal{X}_\gamma$.

The inter-regime variance captures differences across regimes, weighted by their frequency in the top-$\gamma$ region.
However, for conditional hyperparameters, the regime $I^{(d)}$ is determined by other hyperparameters (\eg, parent choices) rather than by the $d$-th hyperparameter itself.
Thus, attributing this inter-regime effect to the $d$-th hyperparameter is often inappropriate.
In fact, we can show that the standard local marginal variance of a conditioned hyperparameter is lower-bounded by the conditioning variable's local marginal variance, implying systematic leakage (\cf \cref{thm:gating-leakage} in \cref{apx:leakage-of-gating-effects-in-standard-local-hpi}). Consequently, even inactive hyperparameters can receive non-negligible importance solely due to upstream choices.
Such behavior is undesirable because tuning an inactive hyperparameter has no effect on the objective and is therefore meaningless.

Motivated by this consideration, we define the local HPI for conditional
hyperparameters using only the within-regime component.
\begin{definition}[Conditional Local HPI]
  For a conditional hyperparameter $x^{(d)}$, we define the \emph{conditional local HPI} at level $\gamma$ as:
  \begin{equation}
    \text{Conditional Local HPI}
    \coloneqq
    v_{\gamma,\mathrm{within}}^{(d)} \Big/ \sum_{d^\prime=1}^{D} v_{\gamma,\mathrm{within}}^{(d^\prime)}\,,
    \label{eq:cond-local-hpi-def}
  \end{equation}
  where $\smash{v_{\gamma,\mathrm{within}}^{(d)}}$ is the within-regime local marginal variance defined as:
  \begin{equation}
    \smash{v_{\gamma,\mathrm{within}}^{(d)}}
    \coloneqq
    \smash{\mathbb{E}_{I^{(d)}} \left[
      \mathrm{Var}_{Z^{(d)}}\left(g_\gamma^{(d)}(I^{(d)},Z^{(d)})\,\middle|\, I^{(d)}\right)
    \right].}
    \label{eq:within-regime-hpi-def}
  \end{equation}
  \label{def:cond-local-hpi}
\end{definition}
Intuitively, $\smash{v_{\gamma,\mathrm{within}}^{(d)}}$ quantifies how much the value of the $d$-th hyperparameter changes the likelihood of being in the tighter top set $\mathcal{X}_{\gamma^\prime}$, after fixing the regime.
When the hyperparameter is not conditional (\ie, $K^{(d)}=1$), the inter-regime term in \cref{eq:var-decomp-within-inter} vanishes and $v_{\gamma,\mathrm{within}}^{(d)}$ reduces to the standard local HPI.
The empirical difference in HPI performance between $\smash{v_{\gamma,\mathrm{within}}^{(d)}}$ and the original $\smash{v_{\gamma}^{(d)}}$ is examined in detail in \cref{sec:ablation-study}.

\subsection{\methodname}
\label{sec:c-ped-anova-alg}

We now derive a closed-form \methodname estimator that efficiently computes the within-regime local marginal variance of each hyperparameter.
We adopt a PED-ANOVA-style construction since it enables fast HPI computation while maintaining accurate estimates (\cf runtime comparisons in \cref{apx:runtime-comparison-details}).
The conditional local HPI then follows immediately by normalization, as defined in \cref{eq:cond-local-hpi-def}.
The pseudocode is provided in \cref{alg:cond-ped-anova}.

For the $d$-th hyperparameter, let
$\mathcal{D}_{\gamma^\prime} \!\coloneqq\! \{(\bm x_n,f(\bm x_n))\!\mid\! f(\bm x_n) \!\le f_{\gamma^\prime}\}$
and
$\mathcal{D}_{\gamma} \!\coloneqq\! \{(\bm x_n,f(\bm x_n))\!\mid\! f(\bm x_n) \!\le f_{\gamma}\}$
denote the top-$\gamma^\prime$ and top-$\gamma$ evaluation subsets, respectively.
For each regime $i^{(d)}\!\in\!\{1,\ldots,K^{(d)}\}$, let $\mathcal{D}_{\gamma^\prime,\,i}^{(d)}$ and $\mathcal{D}_{\gamma,\,i}^{(d)}$ be the samples in regime $i^{(d)}$ within $\mathcal{D}_{\gamma^\prime}$ and $\mathcal{D}_{\gamma}$, respectively.
We define the regime probabilities:
\begin{equation}
  \alpha_i^{(d)}\coloneqq \frac{|\mathcal{D}_{\gamma^\prime,\,i}^{(d)}|}{|\mathcal{D}_{\gamma^\prime}|},
  \quad\text{and}\quad
  \beta_i^{(d)} \coloneqq \frac{|\mathcal{D}_{\gamma,\,i}^{(d)}|}{|\mathcal{D}_{\gamma}|}.
\end{equation}
For each regime $i^{(d)}$, let $p_{\gamma^\prime,\,i}^{(d)}$ and $p_{\gamma,\,i}^{(d)}$ denote the one-dimensional PDFs of the $d$-th hyperparameter.
In practice, these PDFs are estimated using KDE from the samples in $\mathcal{D}_{\gamma^\prime,\,i}^{(d)}$ and $\mathcal{D}_{\gamma,\,i}^{(d)}$, respectively.

With these definitions, we obtain the following closed-form estimation for the within-regime local marginal variance.
\begin{theorem}[\methodname]
  \label{thm:cped-anova}
  Let $0<\gamma^\prime<\gamma\le 1$.
  The within-regime local marginal variance for the $d$-th hyperparameter at level $\gamma$, computed using the indicator function $b_{\gamma^\prime} \coloneqq \mathbf{1}\{x\in\mathcal{X}_{\gamma^\prime}\}$, is given by:
  \begin{equation}
    v_{\gamma,\mathrm{within}}^{(d)}
    =
    \left(\frac{\gamma^\prime}{\gamma}\right)^{\!2}
    \sum_{i=1}^{\;K^{(d)}} \frac{\big(\alpha_i^{(d)}\big)^2}{\beta_i^{(d)}}
    D_{\mathrm{PE}}\left(
      p_{\gamma^\prime,\,i}^{(d)} \,\middle\|\, p_{\gamma,\,i}^{(d)}
    \right).
    \label{eq:cped-anova-within}
  \end{equation}
  By normalizing the variance across all hyperparameters as defined in \cref{eq:cond-local-hpi-def}, we obtain the conditional local HPI for the $d$-th hyperparameter.
\end{theorem}

The proof is provided in \cref{apx:c-ped-anova-details}.
\cref{eq:cped-anova-within} aggregates within-regime Pearson divergence using regime-dependent weights $\smash{\big(\alpha_i^{(d)}\big)^2/\beta_i^{(d)}}$.
Intuitively, this weighting emphasizes regimes that are enriched in the tighter top set (\ie, with larger $\smash{\big(\alpha_i^{(d)}\big)^2/\beta_i^{(d)}}$), while attributing importance only to changes \emph{within} each regime.
If a regime corresponds to an inactive configuration (so the value is always $\bot$), then $\smash{p_{\gamma^\prime,\,i}^{(d)}}$ and $\smash{p_{\gamma,\,i}^{(d)}}$ are degenerate at $\bot$ and the corresponding divergence is zero.
When the hyperparameter is not conditional, we have $\smash{K^{(d)}=1}$ and $\smash{\alpha_1^{(d)}=\beta_1^{(d)}=1}$, so \cref{eq:cped-anova-within} reduces to the original PED-ANOVA in \cref{eq:ped-local-hpi}.


\section{Experiments on Synthetic Problems}
\label{sec:experiments-synthetic}

\begin{figure*}[!t]
  \centering
  \includegraphics[width=\textwidth]{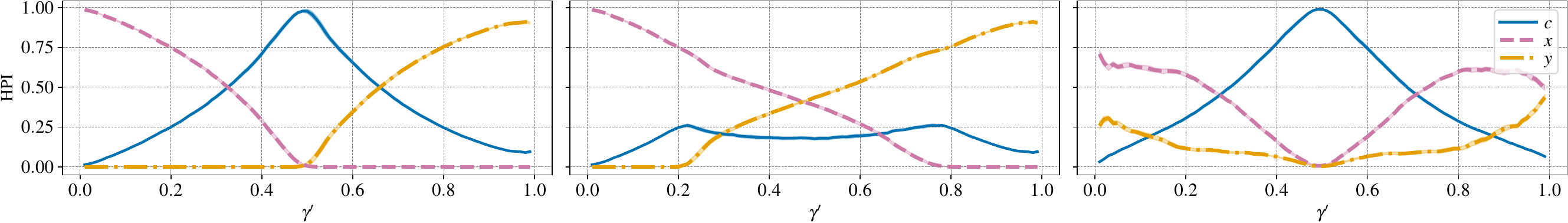}
  \begin{minipage}{0.29\textwidth}
    \centering
    \subcaption{Conditional activation (\cref{eq:toy-objective-conditional}) with disjoint domains (\cref{eq:toy-domains-disjoint})}
    \label{fig:cped-anova-synthetic-objective-results-disjoint}
  \end{minipage}
  \hspace{4mm}
  \begin{minipage}{0.29\textwidth}
    \centering
    \subcaption{Conditional activation (\cref{eq:toy-objective-conditional}) with overlapping domains (\cref{eq:toy-domains-overlap})}
    \label{fig:cped-anova-synthetic-objective-results-overlap}
  \end{minipage}
  \hspace{4mm}
  \begin{minipage}{0.29\textwidth}
    \centering
    \subcaption{Regime-dependent domains (\cref{eq:toy-objective-regime-dependent} and \cref{eq:toy-domain-regime-dependent})}
    \label{fig:cped-anova-synthetic-objective-results-regime-dependent}
  \end{minipage}
  \caption{
    \methodname ($\gamma=1.0$) HPI computed for the synthetic objectives.
    For the objective with conditional activation (\cref{eq:toy-objective-conditional}), the gating hyperparameter $c$ determines which branch is active: $x$ is present only when $c<0.5$, whereas $y$ is present only when $c\ge 0.5$.
    For the objective with regime-dependent domains (\cref{eq:toy-objective-regime-dependent}), the gating hyperparameter $c$ determines the domain of $x$ and $y$.
    The lines denote the mean, and the shaded regions denote the standard error, both computed over ten independent runs with different random seeds.
    The results on other values of $\gamma$ are provided in \cref{fig:cped-anova-synthetic-objective-results-gammas}.
  }
  \label{fig:cped-anova-synthetic-objective-results}
\end{figure*}

Here, we first empirically validate \methodname on synthetic problems to verify that it properly reflects conditional structure, including parameter activation and regime-dependent domains.
We choose a synthetic setting because the expected behavior is clear and easy to diagnose, unlike real-world objectives where the ``correct'' importance patterns are often ambiguous.
After that, we confirm that \methodname also properly works on real-world benchmarks, which is presented in \cref{sec:experiments-real-world}.
Further experimental results, such as runtime comparison and sensitivity analyses with respect to $N$, are available in \cref{apx:additional-results-synthetic}.

\subsection{Experimental Setup}
\label{sec:common-setup}

We use a fixed evaluation budget of $N=1{,}000$ and draw each sample independently from a uniform distribution over the search space.
We consider three choices of $\gamma \in \{1.0,\,0.75,\,0.5\}$.
For each choice of $\gamma$, we sweep $\gamma^\prime$ from $0.01$ up to $\gamma - 0.01$ in increments of $0.01$, and run experiments for all resulting $(\gamma^\prime,\gamma)$ quantile pairs.
Our \methodname implementation is built on top of the PED-ANOVA implementation in
Optuna~\citep{akiba2019optuna}.
We repeat the HPI computation with 10 different random seeds and report the mean and the standard error across runs.
To ensure reproducibility, we publicly release our \methodname implementation and all experimental codes\footnote{\CondPedAnovaGitHubLink}.
Further details are provided in \cref{apx:details-for-comon-experimental-setup}.

\subsection{Conditional Activation (Presence/Absence)}
\label{sec:conditional-presence}

\subsubsection{Problem Setting}
\label{sec:conditional-activation-setting}

We first consider a simple synthetic problem with a conditional search space, where the active variable is selected by a gating parameter.
The objective function is defined as:
\begin{equation}
  f(c,x,y)
  =
  \begin{cases}
    x, & \text{if } c < 0.5,\\
    y, & \text{if } c \ge 0.5,
  \end{cases}
  \quad
  \text{with } c\in[0,1].
  \label{eq:toy-objective-conditional}
\end{equation}
Our goal is to minimize $f$.
When $c<0.5$, the parameter $x$ is active and $y$ is inactive; when $c\ge 0.5$, vice versa.

We study the following two domain settings:
\begin{align}
  \text{(Disjoint domains)}\quad
  &x \in [-5,-2],~y \in [2,5],
  \label{eq:toy-domains-disjoint}\\
  \text{(Overlapping domains)}\quad
  &x \in [-5,2],~y \in [-2,5].
  \label{eq:toy-domains-overlap}
\end{align}
The first setting cleanly separates the two branches, whereas the second introduces overlap, making the conditional structure less trivially identifiable from the objective
values.
Experiments with multi-level conditional hierarchies are presented in \cref{apx:nested-conditional-activation}.

\subsubsection{Results and Discussion}
\label{sec:conditional-activation-results}

The results are shown in \cref{fig:cped-anova-synthetic-objective-results-disjoint,fig:cped-anova-synthetic-objective-results-overlap}.
The value at each $\gamma^\prime$ can be interpreted as a \emph{target-dependent} importance: it indicates which hyperparameters matter most if we aim to optimize the objective up to the top-$\gamma^\prime$ performance.

We first discuss the disjoint-domain setting (\cref{eq:toy-domains-disjoint}) with $\gamma=1.0$ (\cref{fig:cped-anova-synthetic-objective-results-disjoint}).
Around $\gamma^\prime\approx 0.5$, the gating variable $c$ has the largest HPI, whereas for a tighter target (\eg, $\gamma^\prime\approx 0.1$), the HPI shifts to $x$, and for looser targets (\eg, $\gamma^\prime\approx 0.9$), the HPI is dominated by $y$.
This behavior is intuitive.
When $c<0.5$, the objective equals $x$ and lies in a low (better) range; when $c\ge 0.5$, it equals $y$ and lies in a high (worse) range.
Thus, achieving roughly median performance is mainly determined by the selected branch, explaining the high importance of $c$ near $\gamma^\prime\approx 0.5$.
For smaller $\gamma^\prime$, selecting the favorable branch is not enough; one must optimize the active variable, which explains the increasing importance of $x$.
Moreover, for $\gamma^\prime<0.5$, top-$\gamma^\prime$ performance requires the $c<0.5$ branch and $y$ to be inactive, which is consistent with the (near) zero importance of $y$.
Conversely, for $\gamma^\prime>0.5$, the $c\ge 0.5$ branch dominates, making $y$ important and $x$ inactive (hence unimportant).

We next discuss the overlapping-domain setting (\cref{eq:toy-domains-overlap}) with $\gamma=1.0$ (\cref{fig:cped-anova-synthetic-objective-results-overlap}).
Because the domains of $x$ and $y$ overlap, objective values are no longer cleanly separated by the gating variable $c$.
Consequently, around $\gamma^\prime\approx 0.5$ the HPI of $c$ is no longer close to one, and both $x$ and $y$ also have non-negligible HPIs.
As $\gamma^\prime$ decreases, the contribution of $y$ persists until about $\gamma^\prime\approx 0.2$, reflecting the best values attainable in the $y$-branch (near $y=-2$).
For further tighter targets, top-$\gamma^\prime$ performance requires the $c<0.5$ branch, consistent with the (near) zero importance of $y$.
Within this branch, achieving higher performance increasingly depends on the precise value of $x$,
leading to a growing HPI for $x$ as $\gamma^\prime$ decreases.

Also, under both domain settings, the resulting HPI is approximately symmetric for $\gamma^\prime < 0.5$ and $\gamma^\prime > 0.5$, reflecting the objective's top--bottom symmetry.
However, for $\gamma^\prime>0.5$, changing $c$ can still improve the objective, so $c$ retains a non-zero HPI even near $\gamma^\prime\approx 1.0$, reflecting a slight asymmetry caused by minimization.

These results suggest that \methodname yields sensible HPIs for $c$, $x$, and $y$ across target levels $\gamma^\prime$.
The results for varying $\gamma$ are presented and discussed in \cref{apx:cped-anova-synthetic-varying-gamma}, where \methodname consistently shows reasonable behavior.
Additional results under non-uniform sampling are provided in \cref{apx:cped-anova-synthetic-tpe-sampling}, showing that \methodname remains robust beyond uniform sampling.

\begin{figure*}[!b]
  \centering
  \includegraphics[width=\textwidth]{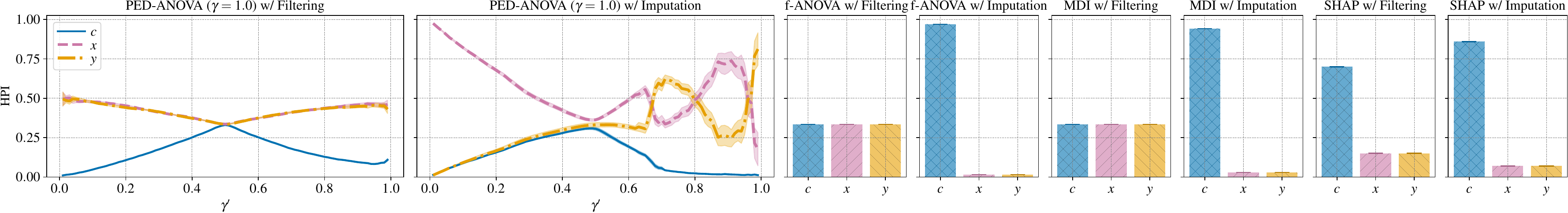}
  \vspace{-6.5mm}
  \caption{
    Baseline HPIs computed with naive extensions of existing methods for the synthetic objective with conditional activation (\cref{eq:toy-objective-conditional}) under the disjoint domain setting (\cref{eq:toy-domains-disjoint}).
    ``Filtering'' computes HPI on the subset of samples where the target hyperparameter is active, whereas ``Imputation'' assigns a default value (the domain midpoint) to inactive samples before computing HPI.
    The lines and bars denote the mean, and the shaded regions and error bars denote the standard error, both computed over ten independent runs with different random seeds. 
  }
  \label{fig:activation-disjoint-results-baseline}
\end{figure*}

\subsection{Regime-Dependent Domains}
\label{sec:regime-dependent-domains}

\subsubsection{Problem Setting}
\label{sec:regime-dependent-domains-setting}

We next consider a synthetic optimization problem where hyperparameter domains change with a gating variable.
The objective is to minimize:
\begin{equation}
  f(c,x,y) = x+y,
  \label{eq:toy-objective-regime-dependent}
\end{equation}
with the following regime-dependent domains:
\begin{equation}
  c\in[0,1],~
  \begin{cases}
    x \in [-7,-2],~y\in [-5,-2], & \text{if } c < 0.5,\\
    x \in [2,7],~y \in [2,5], & \text{if } c \ge 0.5.
  \end{cases}
  \label{eq:toy-domain-regime-dependent}
\end{equation}
Here, both $x$ and $y$ are always active, but their domains switch according to $c$.
Additional experiments combining conditional activation with regime-dependent domains are provided in \cref{apx:three-way-branching}.

\subsubsection{Results and Discussion}
\label{sec:regime-dependent-domains-results}
The results are shown in \cref{fig:cped-anova-synthetic-objective-results-regime-dependent}.
Similar to the disjoint-domain setting (\cref{eq:toy-domains-disjoint}) in the conditional activation experiment (\cref{sec:conditional-presence}),
the gating variable $c$ largely determines top-half performance, yielding high importance near $\gamma^\prime\approx 0.5$.
Unlike that case, however, both $x$ and $y$ are always active and contribute to the objective, consistent with their importance grows toward more extreme targets (smaller or larger $\gamma^\prime$).
Moreover, $x$ is consistently more important than $y$, reflecting its ability to attain more extreme values (around $\pm 7$).

\subsection{Comparison with Existing Methods}
\label{sec:comparison-with-existing-methods}

\subsubsection{Baselines}
\label{sec:baselines}

We then compare \methodname with existing HPI estimators. Since these HPI estimators do not handle conditional hyperparameters in a principled way, we evaluate them using common naive extensions for conditional search spaces.

We consider the following four baselines:
\begin{itemize}
  \item \textbf{PED-ANOVA}~\citep{watanabe2023ped}, the original PED-ANOVA method;
  \item \textbf{f-ANOVA}~\citep{hutter2014efficient}, which estimates HPI via functional ANOVA decomposition on a random forest surrogate;
  \item \textbf{MDI}~\citep{NIPS2013_e3796ae8}, the mean decrease impurity feature importance computed from a random forest surrogate; and
  \item \textbf{SHAP}~\citep{NIPS2017_8a20a862}, which assigns importance based on Shapley values~\citep{Shapley+1953+307+318} computed from a surrogate model, measuring each variable's average marginal contribution.
\end{itemize}

For hyperparameters whose presence changes depending on other variables, we use the following two naive extensions:
\begin{itemize}
  \item \textbf{Filtering:} From the evaluation set $\{(\bm x_n,f(\bm x_n))\}_{n=1}^N$, we keep only samples in which the target hyperparameter is present, and compute HPI on the filtered subset.
  \item \textbf{Imputation:} For samples where the target hyperparameter is inactive, we impute it with a default value (the domain midpoint) and compute HPI on the completed dataset.
\end{itemize}

For hyperparameters whose domain switches across regimes, we use the following naive extension:
\begin{itemize}
  \item \textbf{Expansion:} We treat the hyperparameter as if it were sampled from a single expanded domain covering all regime-specific ranges, and compute HPI on this unified domain.
\end{itemize}

The pseudocodes of these naive extensions are provided in \cref{apx:implementation-details-baselines}.
Consequently, for conditional activation we evaluate 8 baseline variants (4 HPI methods $\times$ 2 handling schemes), while for regime-dependent domains we evaluate 4 variants (4 methods $\times$ 1 expansion scheme), and compare them against \methodname.
See \cref{apx:implementation-details-baselines} for further details on the experimental setup.

\begin{figure*}[!t]
  \centering
  \includegraphics[width=\textwidth]{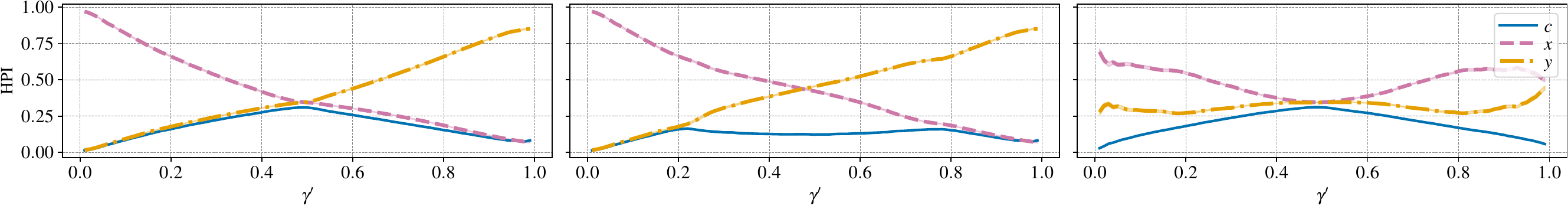}

  \vspace{-4pt}
  \begin{minipage}{0.29\textwidth}
    \centering
    \subcaption{Conditional activation (\cref{eq:toy-objective-conditional}) with disjoint domains (\cref{eq:toy-domains-disjoint})}
    \label{fig:ablation-local-marginal-variance-disjoint}
  \end{minipage}
  \hspace{4mm}
  \begin{minipage}{0.29\textwidth}
    \centering
    \subcaption{Conditional activation (\cref{eq:toy-objective-conditional}) with overlapping domains (\cref{eq:toy-domains-overlap})}
    \label{fig:ablation-local-marginal-variance-overlap}
  \end{minipage}
  \hspace{4mm}
  \begin{minipage}{0.29\textwidth}
    \centering
    \subcaption{Regime-dependent domains (\cref{eq:toy-objective-regime-dependent} and \cref{eq:toy-domain-regime-dependent})}
    \label{fig:ablation-local-marginal-variance-regime-dependent}
  \end{minipage}
  \vspace{-9pt}
  \caption{
    Ablation study results using standard local HPI instead of our conditional local HPI.
    The lines denote the mean, and the shaded regions denote the standard error, both computed over ten independent runs with different random seeds.
  }
  \label{fig:ablation-local-marginal-variance}
\end{figure*}

\begin{figure*}[!t]
  \centering
  \includegraphics[width=\textwidth]{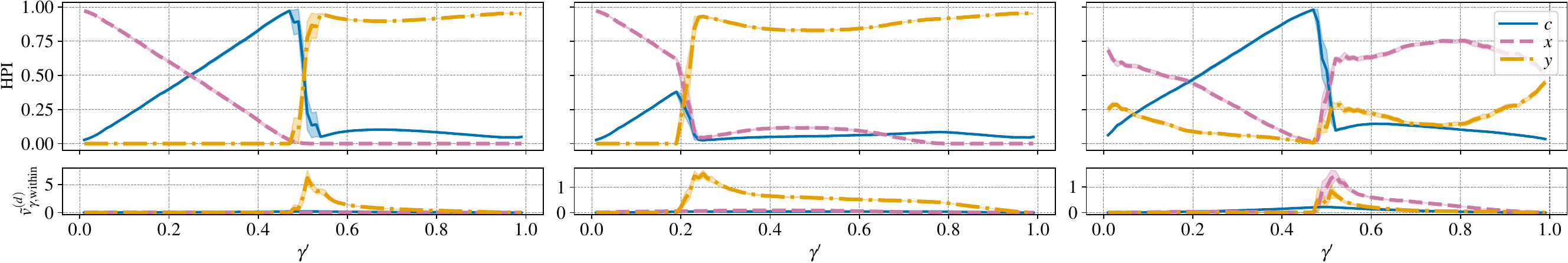}

  \vspace{-4pt}
  \begin{minipage}{0.29\textwidth}
    \centering
    \subcaption{Conditional activation (\cref{eq:toy-objective-conditional}) with disjoint domains (\cref{eq:toy-domains-disjoint})}
    \label{fig:ablation-wo-regime-probabilities-disjoint}
  \end{minipage}
  \hspace{4mm}
  \begin{minipage}{0.29\textwidth}
    \centering
    \subcaption{Conditional activation (\cref{eq:toy-objective-conditional}) with overlapping domains (\cref{eq:toy-domains-overlap})}
    \label{fig:ablation-wo-regime-probabilities-overlap}
  \end{minipage}
  \hspace{4mm}
  \begin{minipage}{0.29\textwidth}
    \centering
    \subcaption{Regime-dependent domains (\cref{eq:toy-objective-regime-dependent} and \cref{eq:toy-domain-regime-dependent})}
    \label{fig:ablation-wo-regime-probabilities-regime-dependent}
  \end{minipage}
  \vspace{-9pt}
  \caption{
    Ablation study results using the naive aggregation scheme (\cref{eq:naive-within-sum}) instead of our expression (\cref{thm:cped-anova}).
    The bottom row shows the HPIs before normalization, \ie, $\smash{\tilde v_{\gamma,\mathrm{within}}^{(d)}}$.
    The lines denote the mean, and the shaded regions denote the standard error, both computed over ten independent runs with different random seeds.
  }
  \label{fig:ablation-wo-regime-probabilities}
\end{figure*}

\subsubsection{Results and Discussion}
\label{sec:comparison-with-existing-methods-results}

We first analyze the results on the conditional activation objective (\cref{eq:toy-objective-conditional}) under the disjoint-domain setting (\cref{eq:toy-domains-disjoint}), which are shown in \cref{fig:activation-disjoint-results-baseline}.
As shown in this figure, applying simple filtering to PED-ANOVA leads to almost identical HPIs for $x$ and $y$.
This behavior is clearly undesirable: in order to achieve better performance (\ie, smaller $\gamma^\prime$), $y$ is inactive, and tuning $y$ cannot affect the objective at all.
This failure arises because filtering obscures how the variable relates to performance over the full search space.
When imputation is applied instead, the HPIs of $x$ and $y$ exhibit substantial instability for $\gamma^\prime > 0.5$.
These behaviors are caused by forcibly imputing an artificial value for inactive samples, which distorts the sample distribution and introduces spurious dependencies.
Furthermore, under both filtering and imputation, PED-ANOVA fails to identify $c$ as important: around $\gamma^\prime\approx 0.5$, it assigns nearly identical HPIs to all hyperparameters, even though $c$ largely determines whether a sample falls in the top-performing half.

Similar issues are observed for the other baselines.
For f-ANOVA and MDI with filtering, all hyperparameters receive identical HPIs, while all the remaining variants assign identical HPIs to $x$ and $y$.

These results show that naive extensions are insufficient for applying existing HPI estimators to conditional search spaces, often yielding uninterpretable HPIs.
In contrast, \methodname consistently yields meaningful HPIs that reflect the underlying conditional structure (as shown in \cref{sec:conditional-presence}).
Additional results for other objectives and domain settings are deferred to \cref{apx:comparison-existing-methods-additional-results}, where the baselines similarly fail in conditional settings.

\subsection{Ablation Study: Effect of Within-Regime Variance and Regime Weighting}
\label{sec:ablation-study}

In this section, we conduct ablation studies to examine the impact of key design choices in \methodname.

\subsubsection{Effect of Within-Regime Variance for HPI}
\label{sec:ablation-within-regime-variance}

In \cref{sec:within-regime-hpi}, we defined the conditional local HPI using the within-regime local marginal variance $v_{\gamma,\mathrm{within}}^{(d)}$.
We therefore compare the resulting HPI estimates with those obtained using the standard local marginal variance $\smash{v_{\gamma}^{(d)}}$.
The closed-form estimation for the standard local marginal variance, computed using the indicator function $b_{\gamma^\prime} \coloneqq \mathbf{1}\{x\in\mathcal{X}_{\gamma^\prime}\}$, is given as follows, where we can also derive a decomposition of the Pearson divergence into within-regime and inter-regime components (\cf \cref{thm:ped-regime} in \cref{apx:ped-regime-decomp}):
\begin{equation}
  \!\!v_{\gamma}^{(d)}
  \!=
  \left(
    \frac{\gamma^\prime}{\gamma}
  \right)^{\!2}
  \!\Bigg(
    \underbrace{
      \!\sum_{i=1}^{\;K^{(d)}}\!\!
      \frac{\big(\alpha_i^{(d)}\big)^{2\!}}{\beta_i^{(d)}}
      D_\mathrm{PE}\!\left(
        p_{\gamma^\prime,\,i}^{(d)}\middle\| p_{\gamma,\,i}^{(d)}\!
      \right)
    }_{\text{within-regime divergence}}
    +
    \hspace{-3pt}
    \underbrace{
      D_\mathrm{PE}\!\left(\bm \alpha^{(d)}\middle\| \bm \beta^{(d)}\right)
    }_{\text{inter-regime divergence}}
    \hspace{-3pt}
  \Bigg)
  ,\!
  \label{eq:standard-local-marginal-variance}
\end{equation}
where $\smash{\bm \alpha^{(d)}} \coloneqq (\smash{\alpha_1^{(d)}}, \ldots, \smash{\alpha_{K^{(d)}}^{(d)}})$ and $\smash{\bm \beta^{(d)}} \coloneqq (\smash{\beta_1^{(d)}}, \ldots, \beta_{K^{(d)}}^{(d)})$.

The results obtained using the standard local marginal variance are shown in \cref{fig:ablation-local-marginal-variance}.
In both the conditional-activation with disjoint domain setting (\cref{fig:ablation-local-marginal-variance-disjoint}) and the regime-dependent domain setting (\cref{fig:ablation-local-marginal-variance-regime-dependent}), the HPI of the gating variable $c$ is underestimated around $\gamma^\prime\approx 0.5$ and even falls below those of $x$ and $y$, despite branch selection being crucial for median-level performance.
Moreover, in the conditional-activation setting (\cref{fig:ablation-local-marginal-variance-disjoint,fig:ablation-local-marginal-variance-overlap}), $x$ and $y$ retain non-zero HPI even where they are inactive (\eg, $y$ for $\gamma^\prime<0.5$ and $x$ for $\gamma^\prime>0.5$ in \cref{fig:ablation-local-marginal-variance-disjoint}).

This happens because the inter-regime variance is primarily induced by the gating variable $c$. When HPI is computed using the standard local marginal variance that includes this inter-regime component, the effect of $c$ leaks into the HPI of $x$ and $y$, suppressing $c$ and assigning spurious HPIs to $x$ and $y$.
In fact, we can theoretically show that, under the standard local marginal variance, the variance of a conditioned hyperparameter includes the gating variance as an additive term, which explains the observed leakage and the resulting misattribution of importance (cf.~\cref{thm:gating-leakage}).
Therefore, in conditional settings, it is essential to compute HPI using the within-regime local marginal variance $\smash{v_{\gamma,\mathrm{within}}^{(d)}}$ rather than the standard local marginal variance $\smash{\smash{v_{\gamma}^{(d)}}}$.

\subsubsection{Effect of Regime-Level Weighting for HPI}
\label{sec:ablation-regime-weighting}

In \cref{sec:c-ped-anova-alg}, we derived a closed-form method to compute $v_{\gamma,\mathrm{within}}^{(d)}$ (\cref{thm:cped-anova}). However, one might wonder whether such a derivation is necessary in practice, and instead consider a naive alternative that simply sums the within-regime Pearson divergences, \ie, using:
\begin{equation}
  \tilde v_{\gamma,\mathrm{within}}^{(d)}
  \coloneqq
  \left(\frac{\gamma^\prime}{\gamma}\right)^{\!2}
  \sum_{i=1}^{K^{(d)}}
  D_{\mathrm{PE}}\!\left(
    p_{\gamma^\prime,\,i}^{(d)} \,\middle\|\, p_{\gamma,\,i}^{(d)}
  \right)
  \label{eq:naive-within-sum}
\end{equation}
in place of \cref{eq:cped-anova-within}. 

\begin{figure*}[!t]
  \centering
  \hspace*{-8mm}\includegraphics[width=\textwidth]{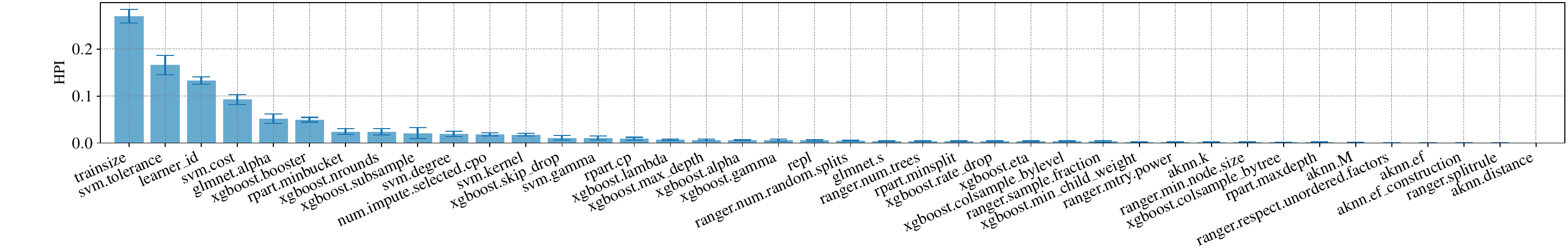}
  \caption{
    \methodname HPIs on a real-world problem from YAHPO Gym \texttt{rbv2\_super} scenario (instance ID 1053)~\citep{pmlr-v188-pfisterer22a,binder2020collecting}.
    The task is model selection over \texttt{ranger}, \texttt{aknn}, \texttt{svm}, \texttt{xgboost}, \texttt{rpart}, and \texttt{glmnet}: the hyperparameter \texttt{learner\_id} selects the learner, activating only the corresponding learner-specific hyperparameters (prefixed by the learner name).
    The bars denote the mean, and the error bars denote the standard error, both computed over ten independent runs with different random seeds.
  }
  \label{fig:yahpo-rbv2-super-1053-cond-ped-anova}
\end{figure*}

The results obtained using this naive aggregation scheme are shown in \cref{fig:ablation-wo-regime-probabilities}.
The resulting HPI curves become noticeably less stable, with near-discontinuous jumps at regime-switching boundaries.
In particular, for the conditional-activation objective under the overlapping-domain setting (\cref{fig:ablation-wo-regime-probabilities-overlap}), the importance of $y$ spikes immediately when $y$ becomes active, excessively dominating the HPI and suppressing $x$ and $c$, which should still contribute to the objective.
Moreover, although the objective is symmetric between the top  ($\gamma^\prime<0.5$) and bottom ($\gamma^\prime>0.5$) region across all settings, the naive scheme breaks this symmetry too severely, yielding clearly distorted importance patterns.

This instability is driven by abrupt scale changes at regime boundaries.
As shown in the bottom row of \cref{fig:ablation-wo-regime-probabilities}, $\tilde v_{\gamma,\mathrm{within}}^{(d)}$ for $x$ and $y$ exhibit implausible jumps at these boundaries, in contrast to the regime-probability-weighted \methodname (\cf \cref{fig:cped-anova-synthetic-objective-results-unnormalized} in \cref{apx:ablation-study-regime-probabilities}).
This failure occurs because removing the regime-probability ratios destroys the frequency-aware weighting, leading to an inappropriate aggregation that ignores regime prevalence.

\section{Experiments on Real-World Benchmarks}
\label{sec:experiments-real-world}


\subsection{Experimental Setup}

We validate that \methodname works properly in real-world settings using YAHPO Gym~\citep{pmlr-v188-pfisterer22a}.
We use the \texttt{rbv2\_super} scenario~\citep{binder2020collecting}, as it defines a conditional search space in a single-objective setting.
This benchmark represents a model selection problem over six learners: \texttt{aknn} (approximate $k$NN), \texttt{glmnet} (regularized linear model), \texttt{ranger} (random forest), \texttt{rpart} (decision tree), \texttt{svm} (support vector machine), and \texttt{xgboost} (gradient-boosted trees).
A categorical hyperparameter \texttt{learner\_id} selects the learner, and only the corresponding learner-specific hyperparameters are active under that choice.
We use a sample size of $N=1000$ and maximize the accuracy (\texttt{acc}) as the objective.
We use $\gamma=1.0$ and $\gamma^\prime=0.1$.
Further details of the experimental setup, including details of the YAHPO Gym, are provided in \cref{apx:detailed-setup-experiments-real-world}.

\begin{table}[t]
  \centering
  \caption{
    Summary statistics of the objective value (\texttt{acc}) for each \texttt{learner\_id} in \texttt{rbv2\_super} (instance ID 1053).
    Values are reported as mean $\pm$ standard error, both computed over 10 independent runs with different random seeds. The \texttt{learner\_id}s are sorted from best to worst max performance.
  }
  \label{tbl:rbv2-super-acc-by-learner}
    \begin{tabular}{lccc}
      \toprule
      \textbf{Learner ID} & \textbf{Min} & \textbf{Mean} & \textbf{Max} \\
      \midrule
      \texttt{svm}     & $0.727 \pm 0.004$ & $0.796 \pm 0.001$ & $0.926 \pm 0.005$ \\
      \texttt{xgboost} & $0.591 \pm 0.011$ & $0.765 \pm 0.001$ & $0.853 \pm 0.004$ \\
      \texttt{rpart}   & $0.736 \pm 0.001$ & $0.774 \pm 0.001$ & $0.846 \pm 0.003$ \\
      \texttt{glmnet}  & $0.731 \pm 0.001$ & $0.778 \pm 0.001$ & $0.842 \pm 0.003$ \\
      \texttt{ranger}  & $0.738 \pm 0.002$ & $0.777 \pm 0.000$ & $0.815 \pm 0.002$ \\
      \texttt{aknn}    & $0.174 \pm 0.001$ & $0.591 \pm 0.006$ & $0.809 \pm 0.002$ \\
      \bottomrule
    \end{tabular}
\end{table}

\subsection{\methodname Results and Discussion}

The \methodname results for the instance ID 1053 are shown in \cref{fig:yahpo-rbv2-super-1053-cond-ped-anova}.
We observe that \texttt{trainsize} and \texttt{learner\_id}, as well as \texttt{svm}- and \texttt{xgboost}-related hyperparameters, have relatively large HPIs, indicating that these factors are most influential.
For reference, we report summary statistics of the objective values (\texttt{acc}) for each selected learner in \cref{tbl:rbv2-super-acc-by-learner}.
The table shows large differences in attainable performance across learners, making it reasonable for larger HPI on \texttt{learner\_id}.
Moreover, the ordering of learners that appear in the computed HPIs (via learner-specific hyperparameters) in \cref{fig:yahpo-rbv2-super-1053-cond-ped-anova} largely matches the learners' performance ordering in \cref{tbl:rbv2-super-acc-by-learner} (\eg, \texttt{svm} higher and \texttt{aknn} lower), suggesting that \methodname correctly captures each learner's relative impact on performance.
Furthermore, within each learner, \methodname assigns high HPIs to hyperparameters that are empirically known to be important, such as \texttt{xgboost}'s \texttt{booster}, \texttt{nrounds} and \texttt{subsample}, and \texttt{svm}'s \texttt{cost}, which is reasonable.
These results indicate that \methodname can effectively identify important hyperparameters in real-world model selection problems with conditional hyperparameters.
Experiments on other problems in YAHPO Gym are reported in \cref{apx:additional-results-real-world-other-instances}, where \methodname constantly produces reasonable HPIs.

\subsection{Comparison with Existing Methods}
\label{sec:experiments-real-world-comparison}

We compare \methodname with naive extensions of the existing HPI estimators considered in \cref{sec:comparison-with-existing-methods}.
We evaluate how well each evaluator captures the differences across learner\_ids. Specifically, we compute the correlation between:
\begin{itemize}
  \item the maximum objective value for each learner\_id (the rightmost column of \cref{tbl:rbv2-super-acc-by-learner}); and
  \item the highest HPI among the hyperparameters associated with each learner\_id produced by each evaluator.
\end{itemize}

\newcommand{\corrtableheaderfont}{\fontsize{7.5}{8.5}\selectfont}

\begin{table*}[t]
  \centering
  \caption{
    Correlation between learner-level performance and learner-specific HPI estimates on \texttt{rbv2\_super}.
    For each instance, we compute the correlation between the maximum objective value for each \texttt{learner\_id} and the highest HPI among the hyperparameters associated with that \texttt{learner\_id}.
  }
  \label{tbl:correlation-with-learner-performance}
  \vspace{-0.2em}
  \setlength{\tabcolsep}{3.5pt}
  \resizebox{\textwidth}{!}{%
    \begin{tabular}{lccccccccc}
      \toprule
      \textbf{Instance ID} &
      \textbf{\shortstack{\corrtableheaderfont condPED-ANOVA\\\corrtableheaderfont (Ours)}} &
      \textbf{\shortstack{\corrtableheaderfont PED-ANOVA\\\corrtableheaderfont w/ Filtering}} &
      \textbf{\shortstack{\corrtableheaderfont PED-ANOVA\\\corrtableheaderfont w/ Imputation}} &
      \textbf{\shortstack{\corrtableheaderfont f-ANOVA\\\corrtableheaderfont w/ Filtering}} &
      \textbf{\shortstack{\corrtableheaderfont f-ANOVA\\\corrtableheaderfont w/ Imputation}} &
      \textbf{\shortstack{\corrtableheaderfont MDI\\\corrtableheaderfont w/ Filtering}} &
      \textbf{\shortstack{\corrtableheaderfont MDI\\\corrtableheaderfont w/ Imputation}} &
      \textbf{\shortstack{\corrtableheaderfont SHAP\\\corrtableheaderfont w/ Filtering}} &
      \textbf{\shortstack{\corrtableheaderfont SHAP\\\corrtableheaderfont w/ Imputation}} \\
      \midrule
      1053 & $\bm{0.98}$ & $0.92$ & $0.93$ & $0.19$ & $-0.46$ & $-0.14$ & $-0.46$ & $-0.44$ & $-0.44$ \\
      1457 & $\bm{0.76}$ & $-0.14$ & $0.61$ & $-0.60$ & $0.70$ & $-0.38$ & $0.60$ & $0.50$ & $0.51$ \\
      1063 & $\bm{0.62}$ & $0.22$ & $0.51$ & $-0.08$ & $-0.72$ & $-0.45$ & $-0.71$ & $-0.65$ & $-0.59$ \\
      1479 & $0.87$ & $-0.03$ & $0.72$ & $0.72$ & $0.90$ & $0.73$ & $0.91$ & $0.96$ & $\bm{0.98}$ \\
      15   & $\bm{0.59}$ & $-0.11$ & $0.54$ & $0.18$ & $0.24$ & $-0.02$ & $0.21$ & $0.14$ & $0.20$ \\
      1468 & $\bm{0.80}$ & $0.13$ & $0.55$ & $-0.27$ & $0.49$ & $-0.19$ & $0.33$ & $0.20$ & $0.30$ \\
      \midrule
      Mean $\pm$ StdErr &
      $\bm{0.77 \pm 0.05}$ &
      $0.16 \pm 0.15$ &
      $0.64 \pm 0.06$ &
      $0.02 \pm 0.17$ &
      $0.19 \pm 0.24$ &
      $-0.07 \pm 0.16$ &
      $0.15 \pm 0.23$ &
      $0.12 \pm 0.22$ &
      $0.16 \pm 0.22$ \\
      \bottomrule
    \end{tabular}%
  }
\end{table*}

The results are shown in \cref{tbl:correlation-with-learner-performance}.
\methodname achieves the highest correlation on almost all instances and does not exhibit the instability observed in baselines, which can yield negative correlations on some instances. This suggests that \methodname best captures differences across learner\_ids.
Experiments on another scenario in YAHPO Gym are reported in \cref{apx:additional-results-real-world-comparison-other-scenario}, where \methodname constantly produces the best correlation without showing instability observed in baselines.

\section{Conclusion}
\label{sec:conclusion}

We studied HPI estimation in conditional search spaces, where hyperparameter activation and domains depend on others.
We identified a key pitfall of standard local HPI: conditioning effects can leak into conditioned hyperparameters, leading to misleading HPIs.
To address this, we introduced conditional local HPI and proposed \methodname, a closed-form, PED-ANOVA-style estimator that correctly handles conditionality while retaining efficiency.
Experiments showed that \methodname produces meaningful HPIs under conditional search spaces, whereas naive extensions of existing estimators often yield misleading HPIs.
We believe \methodname serves as a practical and principled foundation for reliable HPI analysis in HPO workflows with conditional search spaces.


\bibliographystyle{ACM-Reference-Format}
\bibliography{refs}

\appendix

\crefalias{section}{appendix}
\crefalias{subsection}{appendix}
\crefalias{subsubsection}{appendix}

\vspace{5mm}
\section{Additional Related Work}
\label{apx:additional_related_work}

We briefly summarized related work in \cref{sec:related-work}. Here we provide details of representative systems.

\subsection{Contrastive Hyperparameter Importance}
Beyond global variance-based rankings, several lines of work emphasize selection, contrastive explanations, or local sensitivity.
\citet{10.1007/978-3-642-44973-4_40} study \textit{forward selection} to identify a small subset of influential inputs (parameters and instance features) by incrementally adding dimensions that most improve surrogate predictive accuracy, followed by a drop-one analysis to rank selected inputs.
In parallel, \textit{ablation analysis}~\citep{Fawcett2016analysing} explains performance differences between two high-quality configurations by constructing a path that sequentially ablates parameter changes and quantifies their contributions, yielding a contrastive, configuration-to-configuration notion of importance.
\citet{Biedenkapp_Lindauer_Eggensperger_Hutter_Fawcett_Hoos_2017} accelerate this paradigm via \textit{ablation with surrogates}, replacing expensive evaluations with model-based predictions to achieve large speedups while maintaining explanatory utility.
To further focus on neighborhood behavior, CAVE~\citep{10.1007/978-3-030-05348-2_10} introduces local parameter importance, which measures variance in predicted performance when varying one parameter around a specific configuration, and empirically argues that local and global importance can disagree and complement each other.

\subsection{Importance Across Datasets and Tunability}
Another line of work studies the importance beyond a single dataset or run. \citet{van2018hyperparameter} aggregates evidence across many datasets (via OpenML metadata) to identify which hyperparameters tend to matter most for an algorithm in general, and to infer priors over promising values; this reframes HPI as a meta-learning problem and supports more data-driven space design and warm-starting.
Complementarily, \citep{JMLR:v20:18-444} introduce tunability measures that quantify how much performance can improve by tuning, providing a practitioner-facing notion of ``which hyperparameters are worth tuning'' and offering data-based defaults and empirical assessments across datasets and algorithms.

\subsection{Explainable HPO via Partial Dependence and Symbolic Models}
In parallel, a growing body of work tackles explainability of HPO using tools from interpretable machine learning. \citet{NEURIPS2021_12ced2db} study partial dependence plots (PDPs) for explaining BO-based HPO runs and show that naive PDPs can be biased due to the non-uniform, sequential sampling of BO.
They propose uncertainty-aware PDPs based on the BO surrogate, together with a partitioning of the hyperparameter space into subregions where the surrogate is more reliable, to obtain more trustworthy visual summaries of marginal hyperparameter effects.
\citet{pmlr-v224-segel23a} go one step further and introduce symbolic explanations for HPO: using symbolic regression on meta-data collected from BO runs, they learn compact analytic formulas that approximate the relationship between hyperparameters and performance, thereby providing globally interpretable, human-readable explanations that complement fANOVA-style variance decompositions and PDP-based visualizations.

\subsection{Tree-Based Feature Importances}
A widely used family of importance measures arises from tree ensembles such as random forests~\citep{breiman2001random}.
In particular, the mean decrease impurity (MDI) ranks variables by aggregating their impurity reductions over splits across the ensemble.
MDI is popular due to its simplicity and low computational cost, and has been theoretically characterized for randomized trees as a variance decomposition under asymptotic conditions~\citep{NIPS2013_e3796ae8}.
At the same time, impurity-based importances can exhibit biases (\eg, with respect to variable scale or cardinality), motivating careful interpretation and alternative corrections~\citep{strobl2007bias}.

\subsection{Game-Theoretic Explanations for HPI}

Game-theoretic attribution methods provide another approach to importance.
The Shapley value~\citep{Shapley+1953+307+318} offers an axiomatic way to assign each feature its average marginal contribution to a model output.
In machine learning, SHAP~\citep{NIPS2017_8a20a862} leverages Shapley values to produce feature attributions for individual predictions, and global importance scores can be obtained by aggregating these local attributions across data.
These Shapley-based scores are increasingly used as surrogate-based importance baselines in interpretability.

Within HPO, \citet{10.1007/978-3-662-72243-5_30} propose ShapleyBO, explaining Bayesian optimization proposals by attributing each parameter's contribution to the acquisition function rather than to model predictions, and further decomposing contributions into exploration \vs exploitation terms for additive acquisition functions.
More directly for HPI, HyperSHAP~\citep{Wever2026hyperSHAP} uses Shapley values and Shapley interaction indices to provide additive decompositions of a performance measure over hyperparameters, aiming to support both local and global explanations, including interactions.

\subsection{Multi-Objective HPI}
For multi-objective HPO, \citet{Theodorakopoulos_2024} propose a framework that computes HPI across objective trade-offs by applying surrogate-based importance measures (notably f-ANOVA and ablation paths) to scalarized objectives, thereby producing importance profiles that vary with the weighting of competing objectives.

\vspace{1mm}

\section{Rigorous Details of \methodname}
\label{apx:c-ped-anova-details}

We briefly presented an overview of \methodname in \cref{sec:c-ped-anova}.
Here, we provide the rigorous details of the method.

For simplicity, we omit the boldface notation for vectors $x$.
We assume that the objective function $f$ is measurable.

\subsection{Extended One-Dimensional Domain}
\label{apx:extended-domain}

\cref{sec:regime-based-representation} introduced a regime-based representation of conditional hyperparameters.
Here, we provide a rigorous formulation of the corresponding hyperparameter search space.
This formalization is necessary to define the empirical distributions $\mu_{\gamma^\prime}^{(d)}$ and $\mu_\gamma^{(d)}$ (and corresponding PDFs $p_{\gamma^\prime}$ and $p_\gamma$) over a conditional search space, which underlies the subsequent derivations.

We assume that the $d$-th hyperparameter can be in one of $K^{(d)}\in\mathbb{N}$ regimes.
Formally, let
$
  r^{(d)}:\mathcal{X}\to\{1,\ldots,K^{(d)}\}
$
be a measurable regime function that specifies the regime of the $d$-th hyperparameter for each configuration $x\in\mathcal{X}$.
Each regime $i\in\{1,\ldots,K^{(d)}\}$ is associated with its own domain
$\mathcal{Z}_i^{(d)}$, which may differ across regimes.

To capture both regime information and regime-specific domain in a single one-dimensional object, we introduce an \textit{extended domain} for the $d$-th hyperparameter as:
\begin{equation}
  \mathsf{S}^{(d)}
  \coloneqq
  \bigsqcup_{i=1}^{\;\;\;K^{(d)}} \left(\{i\}\times\mathcal{Z}_i^{(d)}\right),
\end{equation}
where $\bigsqcup$ denotes a disjoint union, so that different regimes are treated as separate components even if their numeric domains may overlap.
We equip $\mathsf S^{(d)}$ with the $\sigma$-algebra induced by the disjoint union, so that each partition element $A_i^{(d)} \coloneqq \{i\}\times\mathcal Z_i^{(d)} \subset \mathsf{S}^{(d)}$ is measurable.

\subsection{Rigorous Formulation of \methodname}
\label{apx:c-ped-anova-rigorous}

Let $\mu_{\gamma^\prime}^{(d)}$ and $\mu_\gamma^{(d)}$ denote the empirical
distributions of the extended $d$-th hyperparameter on $\mathsf{S}^{(d)}$
induced by the top-$\gamma^\prime$ and top-$\gamma$ subsets, respectively.
These correspond to densities $p_{\gamma^\prime}$ and $p_\gamma$ introduced in \cref{sec:c-ped-anova} with respect to a suitable reference measure.

For each regime $i$, recall the measurable component
$A_i^{(d)} \subset \mathsf{S}^{(d)}$ and define the regime probabilities:
\begin{equation}
  \alpha_i^{(d)} \coloneqq \mu_{\gamma^\prime}^{(d)}(A_i^{(d)}),
  \quad\text{and}\quad
  \beta_i^{(d)} \coloneqq \mu_{\gamma}^{(d)}(A_i^{(d)}).
\end{equation}
Whenever $\alpha_i^{(d)}>0$ and $\beta_i^{(d)}>0$, define the conditional (restricted) measures:
\begin{equation}
  \mu_{\gamma^\prime,\,i}^{(d)}
  \coloneqq
  \mu_{\gamma^\prime}^{(d)}(\cdot\mid A_i^{(d)}),
  \quad\text{and}\quad
  \mu_{\gamma,\,i}^{(d)}
  \coloneqq
  \mu_\gamma^{(d)}(\cdot\mid A_i^{(d)}).
\end{equation}
If $\alpha_i^{(d)}=0$, we do not need to define $\mu_{\gamma^\prime,\,i}^{(d)}$ since its contribution vanishes in \cref{eq:c-ped-anova-measure} through the factor $\big(\alpha_i^{(d)}\big)^2$.
Since
$\mathsf{S}^{(d)} = \bigsqcup_{i=1}^{K^{(d)}} A_i^{(d)} = \bigsqcup_{i=1}^{K^{(d)}} \big(\{i\}\times\mathcal Z_i^{(d)}\big)$
forms a disjoint partition, we obtain the following regime-wise decomposition of the marginals:
\begin{equation}
  \mu_{\gamma^\prime}^{(d)} = \sum_{i=1}^{\;K^{(d)}} \alpha_i^{(d)}\mu_{\gamma^\prime,\,i}^{(d)}\,,
  \quad\text{and}\quad
  \mu_{\gamma}^{(d)}  = \sum_{i=1}^{\;K^{(d)}} \beta_i^{(d)}\mu_{\gamma,\,i}^{(d)}.
  \label{eq:measure-decomp-regime}
\end{equation}
Note that we have the absolute continuity
$\mu_{\gamma^\prime}^{(d)} \ll \mu_\gamma^{(d)}$, since $\mathcal{X}_{\gamma^\prime}\subset \mathcal{X}_\gamma$.
The regime-wise $\mu_{\gamma^\prime,\,i}^{(d)} \ll \mu_{\gamma,\,i}^{(d)}$ also holds for all regimes with $\beta_i>0$,
since absolute continuity is preserved under restriction to measurable subsets.

With these definitions, we obtain the following closed-form estimation for the within-regime local marginal variance.
\begin{theorem}[\methodname (Restated)]
  \label{thm:c-ped-anova-restated}
  Let $0<\gamma^\prime<\gamma\le 1$.
  The within-regime local marginal variance for the $d$-th hyperparameter at level $\gamma$, computed using the indicator function $b_{\gamma^\prime} \coloneqq \mathbf{1}\{x\in\mathcal{X}_{\gamma^\prime}\}$, is given by:
  \begin{equation}
    v_{\gamma,\mathrm{within}}^{(d)}
    =
    \left(\frac{\gamma^\prime}{\gamma}\right)^{\!2}
    \sum_{i:\,\beta_i^{(d)}>0}\frac{\big(\alpha_i^{(d)}\big)^2}{\beta_i^{(d)}}\,
    D_{\mathrm{PE}}\!\left(
      \mu_{\gamma^\prime,i}^{(d)} \middle\| \mu_{\gamma,i}^{(d)}
    \right),
    \label{eq:c-ped-anova-measure}
  \end{equation}
  By normalizing the variance across all hyperparameters as defined in \cref{eq:cond-local-hpi-def}, we obtain the conditional local HPI for the $d$-th hyperparameter.
\end{theorem}

Here, the Pearson ($\chi^2$) divergence is defined as:
\begin{equation}
  D_{\mathrm{PE}}(\nu\|\mu)
  \coloneqq
  \int \left(\frac{d\nu}{d\mu}-1\right)^{\!2}\, d\mu,
\end{equation}
for probability measures $\nu$ and $\mu$ with $\nu\ll \mu$.
Note that the summation in \cref{eq:c-ped-anova-measure} is restricted to regimes with $\beta_i^{(d)}>0$; regimes absent from the top-$\gamma$ set contribute neither to the within-regime variance nor to the divergence.
If $A_i^{(d)}$ corresponds to an inactive configuration (\ie, $\mathcal{Z}_i^{(d)}=\{\bot\}$), then $\mu_{\gamma,\,i}^{(d)}$ and $\mu_{\gamma^\prime,\,i}^{(d)}$ are both degenerate on $A_i^{(d)}$, implying $D_{\mathrm{PE}}(\mu_{\gamma^\prime,\,i}^{(d)}\|\mu_{\gamma,\,i}^{(d)})=0$.
When the hyperparameter is not conditional, we have $K^{(d)}=1$ and $\alpha_1=\beta_1=1$, so \cref{eq:c-ped-anova-measure}
reduces to the original PED-ANOVA expression.

In practice, the sample size within each regime $A_i^{(d)}$ must be large enough for KDE to be stable.
When the domains $\mathcal{Z}_i^{(d)}$ vary continuously with other coordinates, it is often preferable to discretize them into fixed intervals so that sufficient samples can be pooled within each regime.

\subsection{Proof of \cref{thm:c-ped-anova-restated}}
\label{apx:proof-of-cped-anova}

\begin{proof}

  Let $S$ be a random element distributed according to the empirical distribution
  $\mu_\gamma^{(d)}$ on $\mathsf{S}^{(d)}$ induced by the top-$\gamma$ subset.
  Let $Y\in\{0,1\}$ indicate whether the underlying sample comes from the tighter
  top set, \ie, $Y=1$ corresponds to membership in the top-$\gamma^\prime$ subset.
  By definition, the conditional law of $S$ given $Y=1$ is $\mu_{\gamma^\prime}^{(d)}$.
  Let $\kappa \coloneqq \mathbb{P}(Y=1)$.
  By construction, $\kappa = |\mathcal{D}_{\gamma^\prime}|/|\mathcal{D}_\gamma|$; under the quantile definition used in the main text, this equals $\gamma^\prime/\gamma$ up to the floor effect, and we use $\kappa=\gamma^\prime/\gamma$ for simplicity.

  For any measurable set $B\subset \mathsf{S}^{(d)}$, we have:
  \begin{equation}
    \begin{aligned}
      \mathbb{P}(S\in B,\,Y=1)
      &= \mathbb{P}(Y=1)\,\mathbb{P}(S\in B\mid Y=1) \\
      &= \kappa\,\mu_{\gamma^\prime}^{(d)}(B),
    \end{aligned}
  \end{equation}
  and
  \begin{equation}
    \mathbb{P}(S\in B)=\mu_\gamma^{(d)}(B).
  \end{equation}

  Since $\mathcal{X}_{\gamma^\prime}\subset \mathcal{X}_\gamma$, we have
  $\mu_{\gamma^\prime}^{(d)}\ll \mu_\gamma^{(d)}$, and thus the Radon--Nikod\'ym
  derivative $d\mu_{\gamma^\prime}^{(d)}/d\mu_\gamma^{(d)}$ exists.
  Thus, by the definition of conditional probability, for $s\in \mathsf{S}^{(d)}$, we have:
  \begin{equation}
    \mathbb{P}\!\left(Y=1 \mid S=s\right)
    \;=\;
    \kappa\, \frac{\mu_{\gamma^\prime}^{(d)}}{\mu_{\gamma}^{(d)}}(s).
    \label{eq:condprob-densityratio}
  \end{equation}

  Since $\mathsf{S}^{(d)} = \bigsqcup_{i=1}^{K^{(d)}} A_i^{(d)} = \bigsqcup_{i=1}^{K^{(d)}} \big(\{i\}\times\mathcal Z_i^{(d)}\big)$ forms a disjoint partition (\cref{eq:measure-decomp-regime}), the following factorization holds for each $i$ with $\beta_i^{(d)}>0$, for $s\in A_i^{(d)}$:
  \begin{equation}
    \frac{d\mu_{\gamma^\prime}^{(d)}}{d\mu_{\gamma}^{(d)}}(s)
    =
    \frac{d(\alpha_i^{(d)}\,\mu_{\gamma^\prime,\,i}^{(d)})}{d(\beta_i^{(d)}\,\mu_{\gamma,\,i}^{(d)})}(s)
    =
    \frac{\alpha_i^{(d)}}{\beta_i^{(d)}}\,\frac{d\mu_{\gamma^\prime,\,i}^{(d)}}{d\mu_{\gamma,\,i}^{(d)}}(s),
    \label{eq:rn-factorization}
  \end{equation}

  Using \cref{eq:condprob-densityratio} and \cref{eq:rn-factorization}, for each $i$ with $\beta_i^{(d)}>0$, for $s\in A_i^{(d)}$, we obtain:
  \begin{equation}
    \mathbb{P}(Y=1\mid S=s)
    \;=\;
    \kappa\,\frac{\alpha_i^{(d)}}{\beta_i^{(d)}}\,
    \frac{d\mu_{\gamma^\prime,\,i}^{(d)}}{d\mu_{\gamma,\,i}^{(d)}}(s)
  \end{equation}

  Recall that the within-regime local HPI is defined as the within-regime variance
  of the conditional probability of being in the tighter top set, \ie,
  \begin{equation}
    v_{\gamma,\mathrm{within}}^{(d)}
    \coloneqq
    \mathbb{E}_I\!\left[
      \mathrm{Var}\!\left(\mathbb{P}(Y=1\mid S) \,\middle|\, I\right)
    \right].
  \end{equation}

  Taking expectation over $I$ and using $\mathbb{P}(I=i)=\beta_i$ yields:
  \begin{equation}
    \begin{aligned}
      v_{\gamma,\mathrm{within}}^{(d)}
      &=
      \sum_{i:\,\beta_i^{(d)}>0}
      \beta_i^{(d)}
      \mathrm{Var}_{s\sim \mu_{\gamma,\,i}^{(d)}}\!\left(
        \kappa\frac{\alpha_i^{(d)}}{\beta_i^{(d)}}
        \frac{d\mu_{\gamma^\prime,\,i}^{(d)}}{d\mu_{\gamma,\,i}^{(d)}}(s)
      \right)
      \\
      &=
      \kappa^2
      \sum_{i:\,\beta_i^{(d)}>0}\frac{\big(\alpha_i^{(d)}\big)^2}{\beta_i^{(d)}}
      \mathrm{Var}_{s\sim \mu_{\gamma,\,i}^{(d)}}\!\left(
        \frac{d\mu_{\gamma^\prime,\,i}^{(d)}}{d\mu_{\gamma,\,i}^{(d)}}(s)
      \right)
      \\
      &=
      \kappa^2
      \sum_{i:\,\beta_i^{(d)}>0}\frac{\big(\alpha_i^{(d)}\big)^2}{\beta_i^{(d)}}
      \int \left(\frac{d\mu_{\gamma^\prime,\,i}^{(d)}}{d\mu_{\gamma,\,i}^{(d)}}(s)-1\right)^2 d\mu_{\gamma,\,i}^{(d)}(s)
      \\
      &=
      \kappa^2
      \sum_{i:\,\beta_i^{(d)}>0}\frac{\big(\alpha_i^{(d)}\big)^2}{\beta_i^{(d)}}
      D_{\mathrm{PE}}\!\left(\mu_{\gamma^\prime,\,i}^{(d)} \,\middle\|\, \mu_{\gamma,\,i}^{(d)}\right).
    \end{aligned}
  \end{equation}

  Here, we used the following equation, which holds since $\mu_{\gamma^\prime,i}^{(d)}$ is a probability measure:
  \begin{equation}
    \mathbb{E}_{s\sim \mu_{\gamma,i}^{(d)}}\left[
      \frac{d\mu_{\gamma^\prime,\,i}^{(d)}}{d\mu_{\gamma,\,i}^{(d)}}(s)
    \right]
    =\int \frac{d\mu_{\gamma^\prime,\,i}^{(d)}}{d\mu_{\gamma,\,i}^{(d)}}\,d\mu_{\gamma,\,i}^{(d)}
    =\mu_{\gamma^\prime,\,i}^{(d)}(\mathsf{S}^{(d)})=1.
  \end{equation}

  This completes the proof.
\end{proof}

\section{Detailed Experimental Setup}
\label{apx:experimental-details}

\subsection{Details for Common Experimental Setup}
\label{apx:details-for-comon-experimental-setup}

For estimating the PDFs $p_{\gamma^\prime,\,i}^{(d)}$ and $p_{\gamma,\,i}^{(d)}$, we use KDE with Optuna's default settings, which employ Scott's rule for bandwidth selection~\citep{Scott1992MDE}.
All experiments are run on a single machine with an Intel\textsuperscript{\textregistered} Core\textsuperscript{TM} Ultra 7 155H CPU (22 logical CPUs) and 16\,GB RAM, running Arch Linux.
We use Python version 3.13.11 and Optuna v4.7.0.

\subsection{Implementation Details for Baselines}
\label{apx:implementation-details-baselines}

For all four baseline methods, we use the implementations provided in Optuna~\citep{akiba2019optuna} and follow Optuna's default hyperparameter settings.
The pseudocode for naive extension for these baselines is provided in \cref{alg:naive-filtering,alg:naive-imputation,alg:naive-expansion}

\begin{algorithm}[t]
  \caption{Naive Extension: Filtering for Conditional Presence}
  \label{alg:naive-filtering}
  \begin{algorithmic}[1]
    \Require Evaluation set $\mathcal D = \{(\bm x_n, f(\bm x_n))\}_{n=1}^N$, baseline evaluator $\textsc{BaselineHPI}(\cdot)$
    \Ensure Naive HPI for the hyperparameters $\bm x$ via filtering
    \For{$d=1,\ldots,D$}
      \State $\mathcal{D}_{\mathrm{act}}^{(d)} \gets \{(\bm x_n,f(\bm x_n)) \in \mathcal{D} \mid x_n^{(d)} \text{ is present}\}$
      \Comment filtering
      \If{$|\mathcal{D}_{\mathrm{act}}^{(d)}| = 0$}
        \State $\text{HPI}^{(d)} \gets 0$ \Comment{no active trials; define HPI as zero}
      \Else
        \State $\text{HPI}^{(d)} \gets d\text{-th value of } \textsc{BaselineHPI}(\mathcal{D}_{\mathrm{act}}^{(d)})$
      \EndIf
    \EndFor
    \State \Return $\{\text{HPI}^{(d)}\}_{d=1}^D$
  \end{algorithmic}
  \vspace{4pt}
\end{algorithm}

\begin{algorithm}[t]
  \caption{Naive Extension: Imputation for Conditional Presence}
  \label{alg:naive-imputation}
  \begin{algorithmic}[1]
    \Require Evaluation set $\mathcal D = \{(\bm x_n, f(\bm x_n))\}_{n=1}^N$, hyperparameter domains $\mathcal{X}^{(1)}\times\cdots\times\mathcal X^{(D)}=[\ell^{(1)},u^{(1)}]\times\cdots\times[\ell^{(D)},u^{(D)}]$, baseline evaluator $\textsc{BaselineHPI}(\cdot)$
    \Ensure Naive HPI for the hyperparameters via imputation
    \State Initialize $\tilde{\mathcal{D}} \gets \emptyset$
    \For{$n=1,\ldots,N$}
      \State $\tilde{\bm x}_n \gets \bm x_n$
      \For{$d=1,\ldots,D$}
        \If{$x_n^{(d)} \text{ is inactive}$ in $\bm x_n$}
          \State\Comment imputation with the midpoint of the domain
          \State Set $\tilde{x}_n^{(d)} \gets (\ell^{(d)} + u^{(d)})/2$
        \EndIf
      \EndFor
      \State $\tilde{\mathcal{D}} \gets \tilde{\mathcal{D}} \cup \{(\tilde{\bm x}_n, f(\bm x_n))\}$
    \EndFor
    \State \Return $\textsc{BaselineHPI}(\tilde{\mathcal{D}})$
  \end{algorithmic}
  \vspace{4pt}
\end{algorithm}

\begin{algorithm}[t]
  \caption{Naive Extension: Expansion for Regime-Dependent Domains}
  \label{alg:naive-expansion}
  \begin{algorithmic}[1]
    \Require Evaluation set $\mathcal D = \{(\bm x_n, f(\bm x_n))\}_{n=1}^N$, regime-specific domains for each $d$-th hyperparameter $\{\mathcal{X}^{(d)}_i=[\ell^{(d)}_i,u^{(d)}_i]\}_{i=1}^{K^{(d)}}$, baseline evaluator $\textsc{BaselineHPI}(\cdot)$
    \Ensure Naive HPI for the hyperparameters via domain expansion
    \For{$d=1,\ldots,D$}
      \State $\ell^{(d)}_\mathrm{min} \gets \min_{i \in \{1,\ldots,K^{(d)}\}} \ell^{(d)}_i$
      \State $u^{(d)}_\mathrm{max} \gets \max_{i \in \{1,\ldots,K^{(d)}\}} u^{(d)}_i$
      \State $\mathcal{\tilde X}^{(d)}=[\ell^{(d)}_\mathrm{min},u^{(d)}_\mathrm{max}]$
      \Comment{expanded domain}
    \EndFor
    \State Initialize $\tilde{\mathcal{D}} \gets \emptyset$
    \For{$n=1,\ldots,N$}
      \State $\tilde{\bm x}_n \in \mathcal{\tilde X}^{(d)} \gets \bm x_n$
      \Comment{treat each $x_n^{(d)}$ as a sample from $\mathcal{\tilde X}^{(d)}$}
      \State $\tilde{\mathcal{D}} \gets \tilde{\mathcal{D}} \cup \{(\tilde{\bm x}_n, f(\bm x_n))\}$
    \EndFor
    \State \Return $\textsc{BaselineHPI}(\tilde{\mathcal{D}}, d)$
  \end{algorithmic}
\end{algorithm}

\subsection{Detailed Setup for the Experiments on Real-World Benchmarks (\cref{sec:experiments-real-world})}
\label{apx:detailed-setup-experiments-real-world}

We use YAHPO Gym~\citep{pmlr-v188-pfisterer22a} as a surrogate-based benchmark suite for HPO, which provides fast evaluations for diverse search spaces and datasets.
In YAHPO Gym, a scenario specifies a specific search space and a set of target metrics, while different instances correspond to different datasets/tasks within that scenario.

YAHPO Gym includes the following scenario families:

\begin{itemize}
  \item \textbf{rbv2}: Classical ML pipelines with mixed and often hierarchical/conditional hyperparameter spaces across many OpenML datasets, supporting multi-fidelity via training-size/fraction.
  \item \textbf{nb301}: NAS-Bench-301~\citep{zela2022surrogate} based neural architecture search with a high-dimensional categorical space and epoch-based fidelity, targeting validation accuracy and runtime.
  \item \textbf{lcbench}: Learning-curve benchmark over many OpenML \citep{10.1145/2641190.2641198} tasks with epoch-based fidelity, exposing multiple validation/test performance metrics and training time.
  \item \textbf{iaml}: Interpretable AutoML scenarios that combine predictive performance with resource-usage and interpretability objectives, with multi-fidelity via training-size fractions.
\end{itemize}

Among these, we use the \texttt{rbv2\_super} scenario because it is single-objective and contains conditional search spaces.
We follow the benchmark suites proposed with YAHPO Gym, where instances are selected based on surrogate faithfulness and diversity across scenarios; in the single-objective suite, \texttt{rbv2\_super} includes six instances (which correspond to OpenML dataset IDs): 1053, 1457, 1063, 1479, 15, and 1468.
We conduct experiments on all six instances. We present results for instance ID 1053 in the main text (\cref{sec:experiments-real-world}) and report the remaining results in \cref{apx:additional-results-real-world}.

The remaining experimental settings are the same as those in the synthetic experiments (\cref{sec:common-setup,apx:details-for-comon-experimental-setup}).

\section{Additional Results and Discussion on Experiments on Synthetic Problem (\cref{sec:experiments-synthetic})}
\label{apx:additional-results-synthetic}

\setcounter{figure}{7}
\begin{figure*}[!b]
  \centering
  \includegraphics[width=\textwidth]{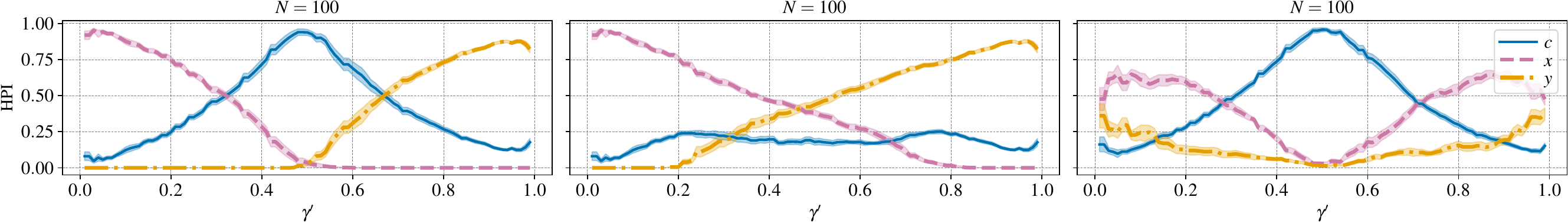}
  \includegraphics[width=\textwidth]{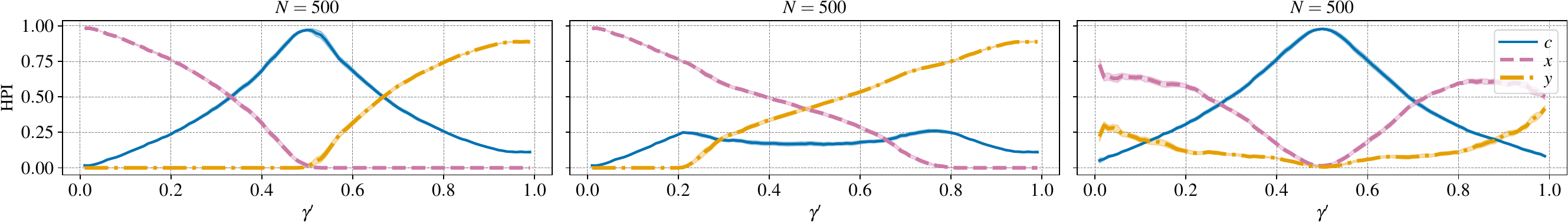}
  \begin{minipage}{0.29\textwidth}
    \centering
    \subcaption{Conditional activation (\cref{eq:toy-objective-conditional}) with disjoint domains (\cref{eq:toy-domains-disjoint})}
    \label{fig:cped-anova-synthetic-objective-results-disjoint-varying-n}
  \end{minipage}
  \hspace{4mm}
  \begin{minipage}{0.29\textwidth}
    \centering
    \subcaption{Conditional activation (\cref{eq:toy-objective-conditional}) with overlapping domains (\cref{eq:toy-domains-overlap})}
    \label{fig:cped-anova-synthetic-objective-results-overlap-varying-n}
  \end{minipage}
  \hspace{4mm}
  \begin{minipage}{0.29\textwidth}
    \centering
    \subcaption{Regime-dependent domains (\cref{eq:toy-objective-regime-dependent} and \cref{eq:toy-domain-regime-dependent})}
    \label{fig:cped-anova-synthetic-objective-results-regime-dependent-varying-n}
  \end{minipage}
  \caption{
    \methodname ($\gamma=1.0$) HPI computed for the synthetic objectives with varying $N$.
    The lines denote the mean, and the shaded regions denote the standard error, both computed over ten independent runs with different random seeds.
  }
  \label{fig:cped-anova-synthetic-objective-results-varying-n}
\end{figure*}

In \cref{sec:experiments-synthetic}, we presented the main results and discussion of experiments on synthetic objectives.
In this section, we provide additional results and discussion that complement those in \cref{sec:experiments-synthetic}.

\subsection{Runtime Comparison}
\label{apx:runtime-comparison-details}

We compare the runtime of \methodname against baseline HPI methods on the synthetic objective with regime-dependent domains (\cref{eq:toy-objective-regime-dependent} and \cref{eq:toy-domain-regime-dependent}).
For each sample size from $2^9$ to $2^{17}$, we measure the wall-clock time required to compute HPI, and report the results in \cref{fig:runtime-comparison}.
f-ANOVA is reported only up to $2^{11}$, since it could not be executed at larger sample sizes due to excessive memory consumption.
Similarly, SHAP is reported only up to $2^{15}$ because it was prohibitively slow at larger sample sizes.

In these results, \methodname runs faster than all baseline methods across the evaluated ranges, indicating that the overhead introduced by our approach is small.
Here, the fact that \methodname is faster than PED-ANOVA indicates that the overhead of filtering or imputation is larger than the additional computation introduced by \methodname.

\setcounter{figure}{5}
\begin{figure}[H]
  \centering
  \hspace{-5mm}\includegraphics[width=0.5\textwidth]{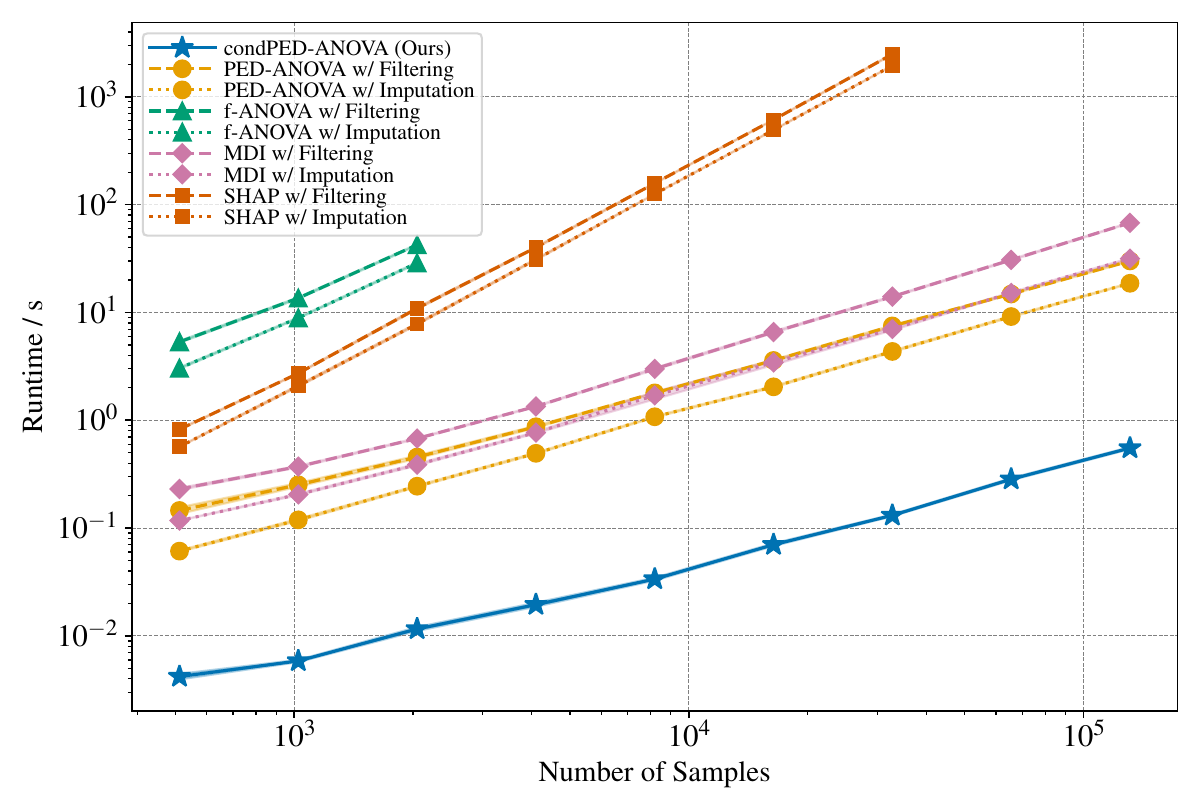}
  \caption{
    Runtime comparison against baseline HPI methods with naive extensions for the synthetic objective (\cref{eq:toy-objective-regime-dependent} and \cref{eq:toy-domain-regime-dependent}).
    The lines denote the mean, and the shaded regions denote the standard error, both computed over ten independent runs with different random seeds. 
  }
  \label{fig:runtime-comparison}
\end{figure}

\subsection{\methodname Results with Different $\gamma$}
\label{apx:cped-anova-synthetic-varying-gamma}

The results of \methodname for the synthetic objectives, with $\gamma$ set to $0.75$ and $0.5$, are shown in \cref{fig:cped-anova-synthetic-objective-results-gammas}.

The results share some similarities with the higher-$\gamma^\prime$ range in \cref{fig:cped-anova-synthetic-objective-results}, which reflects the fact that varying $\gamma$ changes the region over which HPI is computed.
However, the behavior is not a simple zoom-in because the definition of the top-$\gamma$ region itself changes.
In particular, with $\gamma=0.5$ in the disjoint-domain setting (\cref{fig:cped-anova-synthetic-objective-results-overlap-gammas}, right panel), once we restrict attention to the better half of the evaluations (\ie, the region where $c<0.5$ is already selected), adjusting $c$ no longer changes the objective value.
Accordingly, the HPI of $c$ becomes small in this regime, similar to the HPI of $y$.

\setcounter{figure}{6}
\begin{figure}[H]
  \centering
  \begin{minipage}{\linewidth}
    \centering
    \hspace*{-4mm}\includegraphics[width=1.03\linewidth]{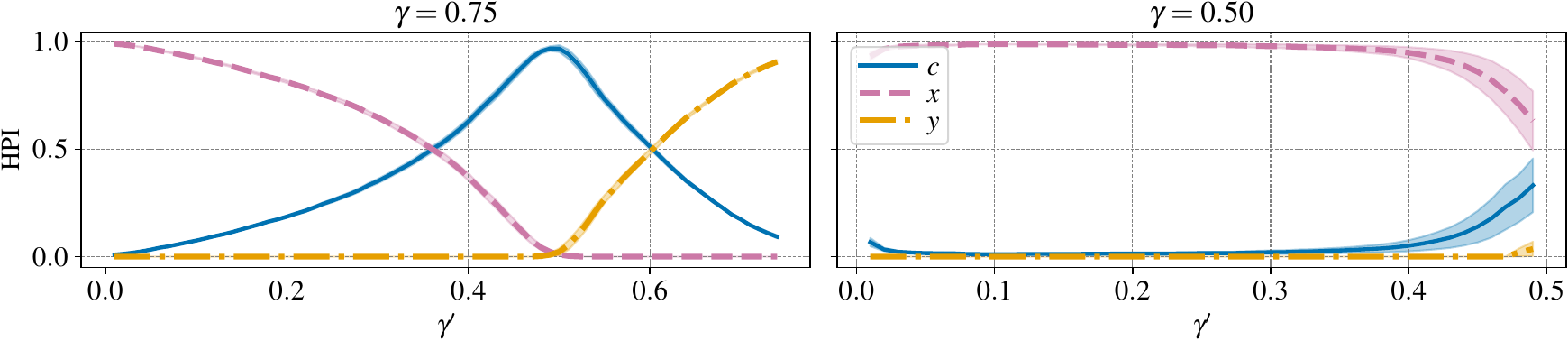}
    \subcaption{Conditional activation (\cref{eq:toy-objective-conditional}) with disjoint domains (\cref{eq:toy-domains-disjoint})}
    \label{fig:cped-anova-synthetic-objective-results-disjoint-gammas}
  \end{minipage}
  \begin{minipage}{\linewidth}
    \centering
    \hspace*{-4mm}\includegraphics[width=1.03\linewidth]{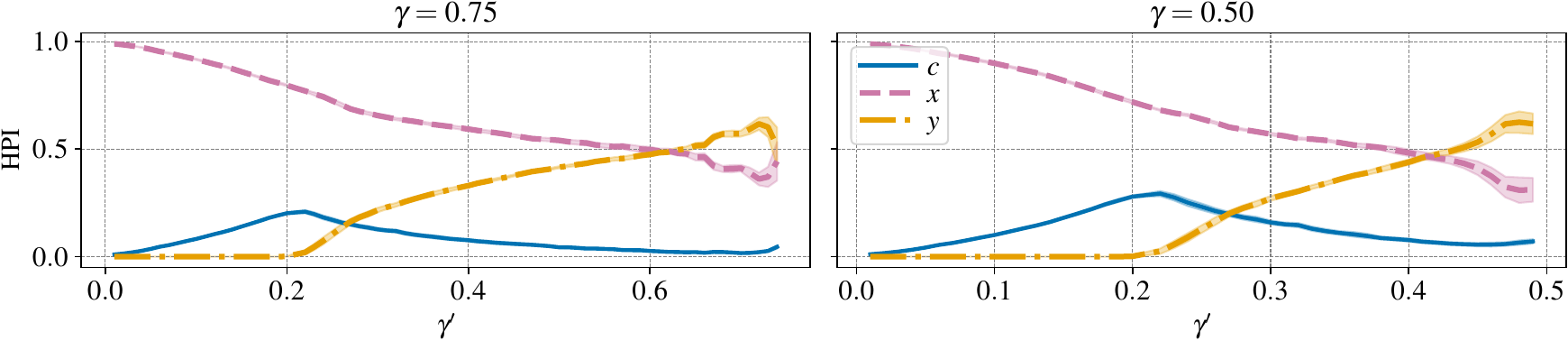}
    \subcaption{Conditional activation (\cref{eq:toy-objective-conditional}) with overlapping domains (\cref{eq:toy-domains-overlap})}
    \label{fig:cped-anova-synthetic-objective-results-overlap-gammas}
  \end{minipage}
  \begin{minipage}{\linewidth}
    \centering
    \hspace*{-4mm}\includegraphics[width=1.03\linewidth]{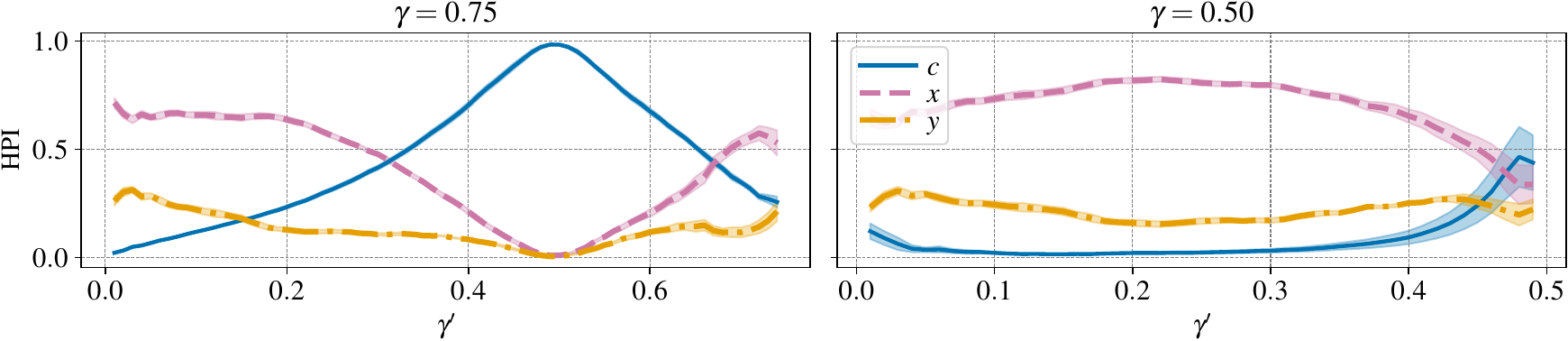}
    \subcaption{Regime-dependent domains (\cref{eq:toy-objective-regime-dependent} and \cref{eq:toy-domain-regime-dependent})}
    \label{fig:cped-anova-synthetic-objective-results-regime-dependent-gammas}
  \end{minipage}
  \caption{
    \methodname HPI computed for the synthetic objectives with varying $\gamma$.
    The lines denote the mean, and the shaded regions denote the standard error, both computed over ten independent runs with different random seeds.
  }
  \label{fig:cped-anova-synthetic-objective-results-gammas}
\end{figure}
\setcounter{figure}{8}

\begin{figure*}[!t]
  \centering
  \includegraphics[width=\textwidth]{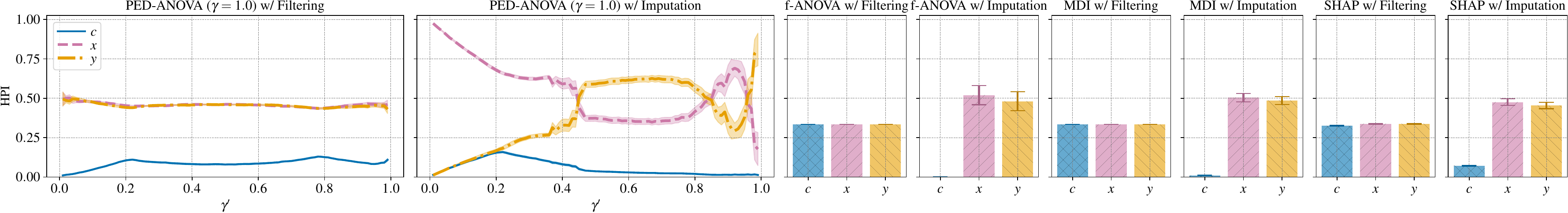}

  \vspace{-3mm}
  \caption{
    Baseline HPIs computed with naive extensions of existing methods for the synthetic objective with conditional activation under the overlapping domain setting (\cref{eq:toy-objective-conditional,eq:toy-domains-overlap}).
    The lines and bars denote the mean, and the shaded regions and error bars denote the standard error, both computed over ten independent runs with different random seeds. 
  }
  \label{fig:activation-overlap-results-baseline}
  \vspace{-1mm}
\end{figure*}

\subsection{\methodname Results with Different $N$}
\label{apx:cped-anova-synthetic-varying-n}

The results of \methodname for the synthetic objectives, with $N$ set to $100$ and $500$, are shown in \cref{fig:cped-anova-synthetic-objective-results-varying-n}.

From this, we observe that reducing $N$ slightly increases the standard error.
However, the resulting HPI estimates remain reasonable and are largely consistent with those obtained when $N=1000$.
These results suggest that \methodname is robust even under smaller evaluation budgets.

\subsection{Comparison with Existing Methods}
\label{apx:comparison-existing-methods-additional-results}

\begin{figure}[!t]
  \centering
  \hspace{-4.3mm}\includegraphics[width=0.5\textwidth]{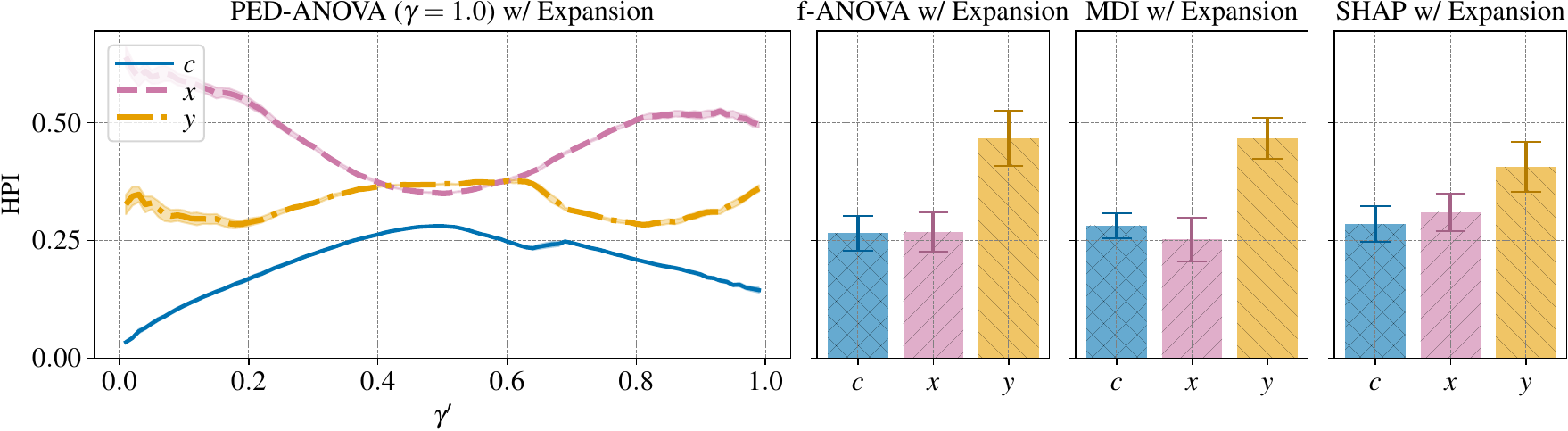}

  \vspace{-3mm}
  \caption{
    Baseline HPIs computed with naive extensions of existing methods for the synthetic objective with regime-dependent domains (\cref{eq:toy-objective-regime-dependent} and \cref{eq:toy-domain-regime-dependent}).
    The lines and bars denote the mean, and the shaded regions and error bars denote the standard error, both computed over ten independent runs with different random seeds. 
  }
  \label{fig:regime-dependent-domains-results-baseline}
  \vspace{-1mm}
\end{figure}

In \cref{sec:comparison-with-existing-methods}, we reported results for baselines computed with naive extensions of existing methods on the conditional-activation objective with the disjoint-domain setting (\cref{eq:toy-objective-conditional,eq:toy-domains-disjoint}).
In this section, we present these baseline results for the conditional-activation objective with the overlapping-domain setting (\cref{eq:toy-objective-conditional,eq:toy-domains-overlap}) and for the regime-dependent domain setting (\cref{eq:toy-objective-regime-dependent,eq:toy-domain-regime-dependent}).
The results are shown in \cref{fig:activation-overlap-results-baseline,fig:regime-dependent-domains-results-baseline}.

\begin{figure*}[!b]
  \centering
  \includegraphics[width=\textwidth]{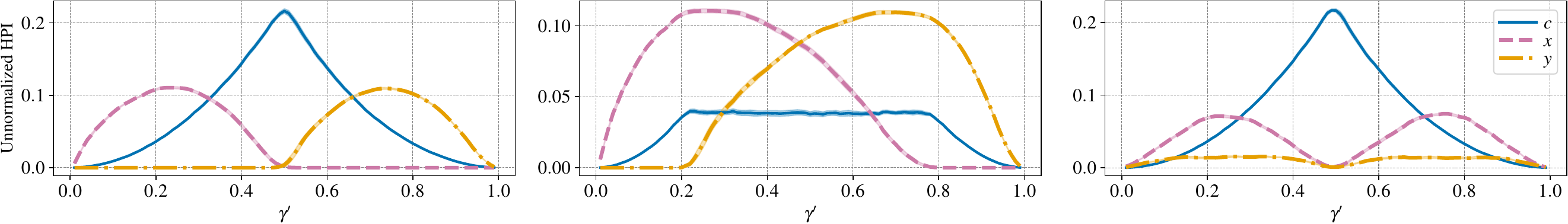}

  \vspace{-2mm}
  \begin{minipage}{0.29\textwidth}
    \centering
    \subcaption{Conditional activation (\cref{eq:toy-objective-conditional}) with disjoint domains (\cref{eq:toy-domains-disjoint})}
    \label{fig:cped-anova-synthetic-objective-results-disjoint-unnormalized}
  \end{minipage}
  \hspace{4mm}
  \begin{minipage}{0.29\textwidth}
    \centering
    \subcaption{Conditional activation (\cref{eq:toy-objective-conditional}) with overlapping domains (\cref{eq:toy-domains-overlap})}
    \label{fig:cped-anova-synthetic-objective-results-overlap-unnormalized}
  \end{minipage}
  \hspace{4mm}
  \begin{minipage}{0.29\textwidth}
    \centering
    \subcaption{Regime-dependent domains (\cref{eq:toy-objective-regime-dependent} and \cref{eq:toy-domain-regime-dependent})}
    \label{fig:cped-anova-synthetic-objective-results-regime-dependent-unnormalized}
  \end{minipage}

  \vspace{-2mm}
  \caption{
    \methodname ($\gamma=1.0$) HPI values before normalization, \ie, $v_{\gamma,\mathrm{within}}^{(d)}$, computed for the synthetic objectives.
    The lines denote the mean, and the shaded regions denote the standard error, both computed over ten independent runs with different random seeds.
  }
  \label{fig:cped-anova-synthetic-objective-results-unnormalized}
\end{figure*}

First, for the conditional-activation objective, we observe the same trend as in the disjoint-domain setting (\cref{eq:toy-objective-conditional,eq:toy-domains-disjoint}).
PED-ANOVA with filtering fails to distinguish between $x$ and $y$, assigning them nearly identical HPI values across all $\gamma^\prime$.
This is clearly problematic, since the contributions of $x$ and $y$ to the objective should differ substantially.
PED-ANOVA with imputation assigns $y$ an importance comparable to $c$ in the small-$\gamma^\prime$ regime, where $y$ should be inactive and thus cannot improve performance; it also exhibits unnatural instability for $\gamma^\prime>0.5$.
f-ANOVA with filtering and MDI with filtering again yield nearly identical HPI values for all hyperparameters, and the remaining baselines also tend to assign almost the same importance to $x$ and $y$, failing to capture the true difference in their contributions to the objective.

Next, for the regime-dependent domain setting, PED-ANOVA with expansion also behaves undesirably: although $c$ determines whether a configuration falls into the top half, the importance of $c$ around $\gamma^\prime \approx 0.5$ becomes smaller than that of $x$ and $y$. This failure is caused by expansion, which collapses the regime-specific ranges of $x$ and $y$ into a unified domain, preventing accurate estimation.
Moreover, for f-ANOVA, MDI, and SHAP with expansion, $y$ is assigned larger importance than $x$, even though $x$ should contribute more to the objective.
This is because forcibly expanding the domain distorts the relationship between the variables and the objective across regimes, leading to misleading HPI estimates.

These results indicate that the naive extensions do not provide meaningful HPI estimates, not only for the conditional-activation objective under the disjoint-domain setting (as shown in \cref{sec:comparison-with-existing-methods-results}), but also under the overlapping-domain setting and in the regime-dependent domain setting.

\subsection{Ablation Study on the Effect of Regime-Level Weighting}
\label{apx:ablation-study-regime-probabilities}

In \cref{fig:ablation-wo-regime-probabilities}, we showed the behavior of the $\tilde v_{\gamma,\mathrm{within}}^{(d)}$ under the naive aggregation scheme (\cref{eq:naive-within-sum}) (bottom row of \cref{fig:ablation-wo-regime-probabilities}).
For reference, we plot $v_{\gamma,\mathrm{within}}^{(d)}$ used by \methodname (which is normalized to obtain conditional local HPI) in \cref{fig:cped-anova-synthetic-objective-results-unnormalized}.

From this, we see that \methodname applies appropriate regime-probability weighting, which calibrates the scale of $v_{\gamma,\mathrm{within}}^{(d)}$ for each variable and keeps it stable across regime changes without large discontinuities.

\subsection{\methodname Results with Non-Uniform Sampling}
\label{apx:cped-anova-synthetic-tpe-sampling}

\begin{figure*}[!t]
  \centering
  \includegraphics[width=\textwidth]{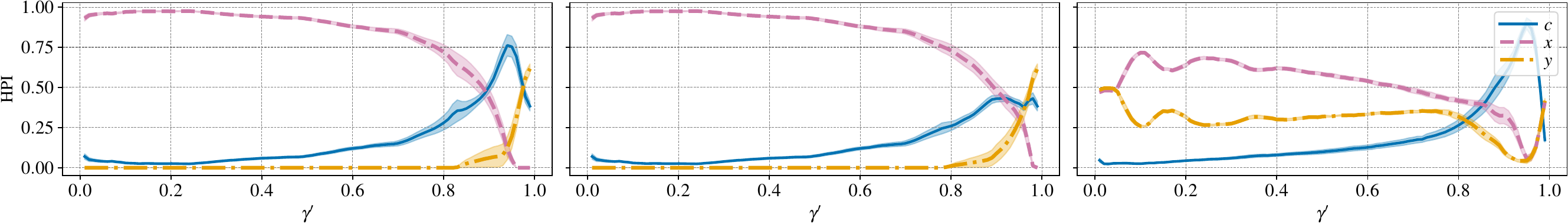}

  \vspace{-2mm}
  \begin{minipage}{0.29\textwidth}
    \centering
    \subcaption{Conditional activation (\cref{eq:toy-objective-conditional}) with disjoint domains (\cref{eq:toy-domains-disjoint})}
    \label{fig:cped-anova-synthetic-objective-results-disjoint-tpe-sampling}
  \end{minipage}
  \hspace{4mm}
  \begin{minipage}{0.29\textwidth}
    \centering
    \subcaption{Conditional activation (\cref{eq:toy-objective-conditional}) with overlapping domains (\cref{eq:toy-domains-overlap})}
    \label{fig:cped-anova-synthetic-objective-results-overlap-tpe-sampling}
  \end{minipage}
  \hspace{4mm}
  \begin{minipage}{0.29\textwidth}
    \centering
    \subcaption{Regime-dependent domains (\cref{eq:toy-objective-regime-dependent} and \cref{eq:toy-domain-regime-dependent})}
    \label{fig:cped-anova-synthetic-objective-results-regime-dependent-tpe-sampling}
  \end{minipage}

  \vspace{-2mm}
  \caption{
    \methodname ($\gamma=1.0$) HPI computed for the synthetic objectives using samples collected by TPE.
    The lines denote the mean, and the shaded regions denote the standard error, both computed over ten independent runs with different random seeds.
  }
  \label{fig:cped-anova-synthetic-objective-results-tpe-sampling}
\end{figure*}

In the experiments based on \cref{sec:conditional-activation-setting}, we used uniformly sampled configurations for simplicity.
Here, we additionally verify that \methodname works properly even under non-uniform sampling.
Specifically, we evaluate \methodname on configurations sampled by the Tree-structured Parzen estimator (TPE)~\citep{NIPS2011_86e8f7ab}, a common Bayesian optimization method.
The results are shown in \cref{fig:cped-anova-synthetic-objective-results-tpe-sampling}.

The results show that \methodname remains robust and consistent with the main results in \cref{sec:conditional-activation-setting}.
Compared with \cref{fig:cped-anova-synthetic-objective-results}, the HPI curves exhibit a rightward shift while largely preserving their overall shapes.
For example, in the conditional-activation objective (\cref{eq:toy-objective-conditional}) with disjoint domains (\cref{eq:toy-domains-disjoint}), the peak around $\gamma^\prime=0.5$ under uniform sampling moves to around $\gamma^\prime=0.9$ under TPE sampling.
This behavior is reasonable and desirable because the HPI reflects the actual sampling distribution used to collect the evaluations, which favors better-performing regions under TPE.
Apart from this shift, the qualitative patterns remain consistent with the uniform-sampling results, demonstrating that \methodname is robust to non-uniform sampling.

\begin{figure}[!b]
  \centering
  \hspace{-5mm}\includegraphics[width=\linewidth]{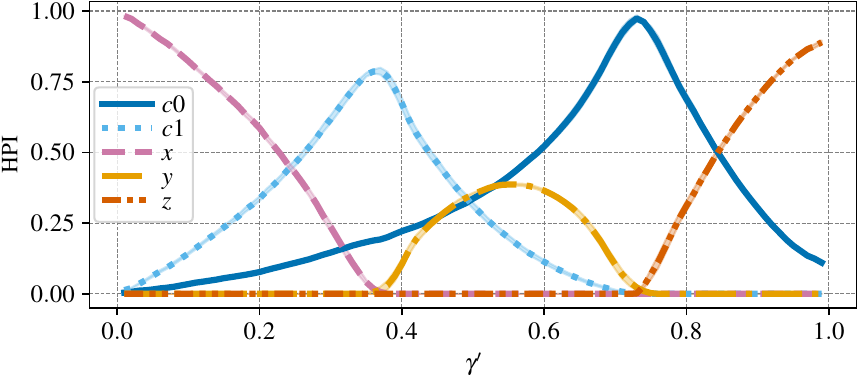}

  \vspace{-2mm}
  \caption{
    \methodname ($\gamma=1.0$) HPI computed for the synthetic objectives on nested conditional activations (\cref{eq:toy-objective-nested-conditional,eq:toy-objective-domains-nested-conditional}).
    The lines denote the mean, and the shaded regions denote the standard error, both computed over ten independent runs with different random seeds.
  }
  \label{fig:cped-anova-synthetic-objective-results-nested-conditions}
\end{figure}

\section{Additional Experiments on Synthetic Problems}
\label{apx:additional-experiments-synthetic}

In \cref{sec:experiments-synthetic}, we considered the simplest synthetic problems. In this section, we study more complex synthetic settings.

\subsection{Nested Conditional Activation}
\label{apx:nested-conditional-activation}

\subsubsection{Problem Setting}
\label{sec:nested-conditional-activation-setting}

We next consider a synthetic problem with a nested conditional search space, where the active variable is selected through a two-level gating hierarchy.
The objective function is defined as:
\begin{equation}
  f(c_0,c_1,x,y,z)
  =
  \begin{cases}
    x, & \text{if } c_0 < 0.75 \text{ and } c_1 < 0.5,\\
    y, & \text{if } c_0 < 0.75 \text{ and } c_1 \ge 0.5,\\
    z, & \text{if } c_0 \ge 0.75,
  \end{cases}
  \label{eq:toy-objective-nested-conditional}
\end{equation}
with the following variable domains:
\begin{equation}
  \begin{aligned}
    c_0 &\in [0,1], \\
    c_1 &\in [0,1],
  \end{aligned}
  \qquad
  \begin{aligned}
    x &\in [-7,-4], \\
    y &\in [-3,0], \\
    z &\in [2,5].
  \end{aligned}
  \label{eq:toy-objective-domains-nested-conditional}
\end{equation}
Our goal is to minimize $f$.
When $c_0<0.75$, a second gating variable $c_1$ becomes active and selects between $x$ (if $c_1<0.5$) and $y$ (if $c_1\ge 0.5$); when $c_0\ge 0.75$, only the variable $z$ is active, and $c_1$, $x$, and $y$ are inactive.

\subsubsection{Results and Discussion}
\label{sec:nested-conditional-activation-results}

The results are shown in \cref{fig:cped-anova-synthetic-objective-results-nested-conditions}.
We observe that $c_0$ attains the highest HPI around $\gamma^\prime\approx 0.75$, taking an importance close to one. This is reasonable, as $c_0$ induces the primary split of the search space. We also see that $c_1$ becomes important around $\gamma^\prime\approx 0.375$. Although its peak is lower than that of $c_0$ (around 0.75), this correctly reflects that, conditional on $c_0<0.75$, $c_1$ provides the next major bifurcation. Moreover, the importance shifts to the active variable within each regime: $x$ becomes large for $\gamma^\prime<0.375$, $y$ for $0.375<\gamma^\prime<0.75$, and $z$ for $\gamma^\prime>0.75$, with their HPI increasing as $\gamma^\prime$ moves away from the corresponding boundaries.
These patterns are consistent with the nested conditional structure, indicating that \methodname yields sensible importances even in nested conditional settings.

\subsection{Combined Conditional Activation and Domain Shifts in a Three-Way Branching}
\label{apx:three-way-branching}

\subsubsection{Problem Setting}
\label{sec:three-way-branching-setting}

We next consider a synthetic problem with a three-way branching conditional search space, where the gating variable induces both conditional activation and regime-dependent domain shifts.
The objective function is defined as:
\begin{equation}
  f(c,x,y)
  =
  \begin{cases}
    x, & \text{if } c < 1/3, \hspace{9.15mm} x \in [-7,-4],\\
    x, & \text{if } 1/3 \le c < 2/3, \hspace{1mm} x \in [-3,0],\\
    y, & \text{if } 2/3 \le c, \hspace{9.4mm} y \in [4,7],
  \end{cases}
  \ c\in[0,1].
  \label{eq:toy-objective-three-way-branching}
\end{equation}
Our goal is to minimize $f$.
Here, $x$ is always active when $c<2/3$, but its domain depends on the regime induced by $c$; when $c\ge 2/3$, $y$ becomes active and $x$ is inactive.

\subsubsection{Results and Discussion}
\label{sec:three-way-branching-results}
\begin{figure}[!t]
  \centering
  \hspace{-5mm}\includegraphics[width=\linewidth]{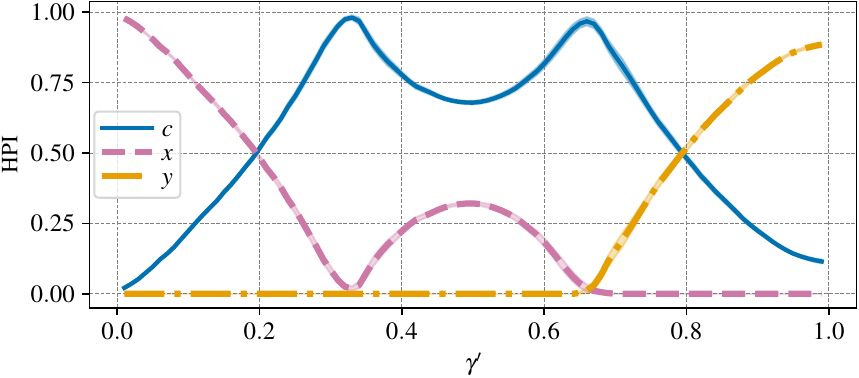}
  \caption{
    \methodname ($\gamma=1.0$) HPI computed for the synthetic objectives on the combined conditional activation and domain shifts in a three-way branching setting (\cref{eq:toy-objective-three-way-branching}).
    The lines denote the mean, and the shaded regions denote the standard error, both computed over ten independent runs with different random seeds.
  }
  \label{fig:cped-anova-synthetic-objective-results-three-way-branching}
\end{figure}

The results are shown in \cref{fig:cped-anova-synthetic-objective-results-three-way-branching}.

We observe that $c$ exhibits pronounced peaks around $\gamma' \approx 0.33$ and $\gamma' \approx 0.66$.
This behavior is reasonable, since it is consistent with the role of $c$ in partitioning the objective-value range into three regimes.
For $x$, the HPI is large in the regions $\gamma' < 0.33$ and $0.33 < \gamma' < 0.66$, while it becomes nearly zero for $\gamma' > 0.66$.
This pattern reflects that $x$ is influential when $c < 1/3$ and when $1/3 \le c < 2/3$.
The near-zero HPI for $\gamma' > 0.66$ is also reasonable, as it corresponds to the regime where $x$ is inactive.
Finally, the HPI of $y$ increases only in the $\gamma' > 0.66$ region and remains close to zero elsewhere.
This is consistent with the fact that $y$ is active only in the regime $c \ge 2/3$.

These results suggest that \methodname correctly computes HPI for branching hyperparameters such as $c$ that induce three or more regimes, as well as for hyperparameters such as $x$ whose activation and domain both vary across regimes.

\section{Supplementary Analysis for the Ablation Study Using Standard Local HPI}
\label{apx:ablation-details}

In this section, we provide additional theoretical details for the ablation study on the effect of within-regime variance for HPI presented in \cref{sec:ablation-within-regime-variance}.
In \cref{apx:leakage-of-gating-effects-in-standard-local-hpi}, we give a theoretical analysis showing that gating effects can leak when within-regime local marginal variance is not used.
In addition, \cref{sec:ablation-within-regime-variance} compared \methodname against a PED-ANOVA-style construction of the standard local HPI using $v_{\gamma}^{(d)}$ as an ablation baseline (\cref{eq:standard-local-marginal-variance}); we provide its derivation in \cref{apx:ped-regime-decomp}.

\subsection{Leakage of Gating Effects in Standard Local HPI}
\label{apx:leakage-of-gating-effects-in-standard-local-hpi}

Let $b_{\gamma^\prime} \coloneqq \mathbf{1}\{\bm x \in \mathcal{X}_{\gamma^\prime}\}$ be the top-set indicator.
Consider a \emph{conditioning} hyperparameter $x^{(c)}$ and a \emph{conditioned} hyperparameter $x^{(d)}$ whose regime is determined by $x^{(c)}$.
Formally, assume that for a conditional hyperparameter $x^{(d)}$ there exists a measurable mapping $\phi^{(d)}\colon \mathcal{X}^{(c)} \to \{1,\ldots,K^{(d)}\}$ such that:
\begin{equation}
  I^{(d)} = \phi^{(d)}(x^{(c)})
  \ \ \text{\as},
  \label{eq:gating-assumption}
\end{equation}
where $I^{(d)}\in\{1,\ldots,K^{(d)}\}$ denotes the random variable representing the regime index of the $d$-th hyperparameter.
Moreover, assume that $x^{(c)}$ affects $b_{\gamma^\prime}$ only through the induced regime (\ie, $x^{(c)}$ has no within-regime effect):
\begin{equation}
  \mathbb{E}_\gamma\left[b_{\gamma^\prime} \mid x^{(c)}\right]
  =
  \mathbb{E}_\gamma\left[b_{\gamma^\prime} \mid I^{(d)}\right]
  \quad \text{\as},
  \label{eq:gating-only-through-regime}
\end{equation}
where the expectations are taken under the empirical distribution restricted to the top-$\gamma$ region $\mathcal{X}_\gamma$.

Here, $x^{(c)}$ corresponds to the gating variable $c$ in the synthetic objectives (\cref{eq:toy-objective-conditional,eq:toy-objective-regime-dependent}), while $x^{(d)}$ corresponds to the conditioned variables $x$ and $y$.
These assumptions are satisfied by the synthetic objectives used in our experiments.

Then, we have the following result regarding the standard local HPI of the conditional hyperparameter $x^{(d)}$.
\begin{theorem}[Leakage of gating effects under the standard local HPI]
\label{thm:gating-leakage}
Under the above assumptions (\cref{eq:gating-assumption,eq:gating-only-through-regime}), the standard local marginal variance of the conditional hyperparameter $x^{(d)}$ contains the local marginal variance of the gating hyperparameter $x^{(c)}$:
\begin{equation}
  v_{\gamma}^{(d)} = v_{\gamma,\mathrm{within}}^{(d)} + v_{\gamma}^{(c)}.
  \label{eq:gating-leakage-decomp}
\end{equation}
In particular, $v_{\gamma}^{(c)} \le v_{\gamma}^{(d)}$.
\end{theorem}

The proof is provided in \cref{apx:proof-theorem-gating-leakage}.
Consequently, when HPI is computed from $v_{\gamma}^{(d)}$, the effect of the gating variable $x^{(c)}$ is additively inherited by conditional hyperparameters through the inter-regime term.

This theorem is consistent with our empirical observations.
In \cref{fig:ablation-local-marginal-variance}, when using the standard local HPI, the importance of the gating variable $c$ is always no larger than those of $x$ and $y$.
Moreover, in the regions of $\gamma^\prime$ where $x$ or $y$ is inactive, their HPI becomes almost identical to that of $c$.
This is exactly what \cref{thm:gating-leakage} predicts: once $v_{\gamma,\mathrm{within}}^{(d)}$ vanishes for an inactive variable, the standard local variance $v_{\gamma}^{(d)}$ reduces to the additive gating term $v_{\gamma}^{(c)}$.

\subsection{Proof of \cref{thm:gating-leakage}}
\label{apx:proof-theorem-gating-leakage}

\begin{proof}

In this section, all expectations and variances are taken under the empirical distribution restricted to the top-$\gamma$ region $\mathcal{X}_\gamma$, and we omit the subscript $\gamma$ on $\mathbb{E}$ and $\mathrm{Var}$ for notational simplicity.

Recall we defined
$
  g_\gamma^{(d)}(I^{(d)},Z^{(d)})
  \coloneqq
  \mathbb{E}\!\left[b_{\gamma^\prime} \mid I^{(d)},Z^{(d)}\right]
$
in \cref{sec:within-regime-hpi}.
We also restate the variance decomposition in \cref{eq:var-decomp-within-inter} here for convenience:
\begin{equation}
  \begin{aligned}
    v_{\gamma}^{(d)}
    &=
    \mathrm{Var}_{I^{(d)},\,Z^{(d)}}\left(g_\gamma^{(d)}(I^{(d)},Z^{(d)})\right)
    \\
    &=
    \underbrace{
      \mathbb{E}_{I^{(d)}}\left[
        \mathrm{Var}_{Z^{(d)}}\left(g_\gamma^{(d)}(I^{(d)},Z^{(d)}) \,\middle|\, I^{(d)}\right)
      \right]
    }_{\text{within-regime variance: } v_{\gamma,\mathrm{within}}^{(d)}}
    \\
    &\hspace{3em}
    +\underbrace{
      \mathrm{Var}_{I^{(d)}}\left(
        \mathbb{E}_{Z^{(d)}}\left[g_\gamma^{(d)}(I^{(d)},Z^{(d)}) \,\middle|\, I^{(d)}\right]
      \right)
    }_{\text{inter-regime variance}}.
    \label{eq:proof-var-decomp}
  \end{aligned}
\end{equation}

By the tower property, we have:
\begin{align}
    \mathbb{E}_{Z^{(d)}}\left[
      g_\gamma^{(d)}(I^{(d)},Z^{(d)}) \,\middle|\, I^{(d)}
    \right] &=
    \mathbb{E}_{Z^{(d)}}\Big[
      \mathbb{E}\big[b_{\gamma^\prime} \,\big|\, I^{(d)}, Z^{(d)}\big]
      \,\Big|\, I^{(d)}
    \Big]
    \notag\\
    &=
    \mathbb{E}\!\left[b_{\gamma^\prime} \mid I^{(d)}\right].
\end{align}
Thus, the inter-regime variance in \cref{eq:proof-var-decomp} equals:
\begin{equation}
  \begin{aligned}
    v_{\gamma,\mathrm{inter}}^{(d)}
    &\coloneqq
    \mathrm{Var}_{I^{(d)}}\left(
      \mathbb{E}_{Z^{(d)}}\left[g_\gamma^{(d)}(I^{(d)},Z^{(d)}) \,\middle|\, I^{(d)}\right]
    \right)\\
    &\,=
    \mathrm{Var}_{I^{(d)}}\!\left(
      \mathbb{E}\left[b_{\gamma^\prime} \mid I^{(d)}\right]
    \right).
    \label{eq:proof-inter-regime-var}
  \end{aligned}
\end{equation}

On the other hand, the local marginal mean of the top-set indicator $b_{\gamma^\prime}$ with respect to the gating coordinate $x^{(c)}$ is:
\begin{equation}
  \begin{aligned}
    f_\gamma^{(c)}(x^{(c)})
    &\coloneqq
    \frac{1}{Z_\gamma^{(c)}(x^{(c)})}
    \sum_{n=1}^N b_{\gamma^\prime}(x^{(c)}, x^{(\bar c\,)}_n)\, b_\gamma(x^{(c)}, x^{(\bar c\,)}_n),
    \\
    &\,=
    \mathbb{E}\left[b_{\gamma^\prime} \mid x^{(c)}\right].
  \end{aligned}
\end{equation}
Thus, by \cref{eq:gating-only-through-regime}, we have:
\begin{equation}
  \begin{aligned}
    v_{\gamma}^{(c)}
    &\coloneqq
    \mathrm{Var}_{x^{(c)}}\left(f_\gamma^{(c)}(x^{(c)})\right)
    \\
    &\,=
    \mathrm{Var}_{x^{(c)}}\left(
      \mathbb{E}\left[b_{\gamma^\prime} \mid x^{(c)}\right]
    \right)
    \\
    &\,=
    \mathrm{Var}_{x^{(c)}}\left(
      \mathbb{E}\left[b_{\gamma^\prime} \mid I^{(d)}\right]
    \right).
  \end{aligned}
  \label{eq:proof-gating-var}
\end{equation}

From \cref{eq:gating-assumption}, the quantity $\mathbb{E}[b_{\gamma^\prime} \mid I^{(d)}]$ is a function of $x^{(c)}$ through $I^{(d)}$.
Therefore,
\begin{equation}
  \mathrm{Var}_{x^{(c)}}\!\left(\mathbb{E}\!\left[b_{\gamma^\prime} \mid I^{(d)}\right]\right)
  =
  \mathrm{Var}_{I^{(d)}}\!\left(\mathbb{E}\!\left[b_{\gamma^\prime} \mid I^{(d)}\right]\right).
  \label{eq:proof-var-eq}
\end{equation}

Comparing \cref{eq:proof-inter-regime-var,eq:proof-gating-var,eq:proof-var-eq}, we conclude that the inter-regime variance of the conditional hyperparameter $x^{(d)}$ equals the marginal variance of the gating hyperparameter $x^{(c)}$:
\begin{equation}
  v_{\gamma,\mathrm{inter}}^{(d)} = v_{\gamma}^{(c)}.
\end{equation}
Substituting this into \cref{eq:proof-var-decomp} completes the proof.
\end{proof}

\subsection{Decomposition of PED over Regimes}
\label{apx:ped-regime-decomp}

Here, we use the notation defined in \cref{apx:extended-domain,apx:c-ped-anova-rigorous}.
That is, on the extended domain $\mathsf S^{(d)}$, $\mu_{\gamma}^{(d)}$ and $\mu_{\gamma^\prime}^{(d)}$ denote the empirical distributions of the extended $d$-th hyperparameter induced by the samples in $\mathcal X_\gamma$ and $\mathcal X_{\gamma^\prime}$, respectively.

We can then show the following decomposition of the Pearson divergence between $\mu_{\gamma}^{(d)}$ and $\mu_{\gamma^\prime}^{(d)}$ on $\mathsf S^{(d)}$.

\begin{lemma}[Regime-wise decomposition of Pearson divergence]
\label[lemma]{thm:ped-regime}
Assume $\mu_{\gamma^\prime}^{(d)} \ll \mu_{\gamma}^{(d)}$ and
$\mu_{\gamma^\prime,\,i}^{(d)} \ll \mu_{\gamma,\,i}^{(d)}$ for every $i$ with $\beta_i^{(d)}>0$.
Then the Pearson divergence between the extended marginals satisfies:
\begin{equation}
  \begin{aligned}
    D_\mathrm{PE}\left(
      \mu_{\gamma^\prime}^{(d)} \,\middle\|\, \mu_{\gamma}^{(d)}
    \right)
    &=
    \underbrace{
      \sum_{i=1}^{\;K^{(d)}}
      \frac{\big(\alpha_i^{(d)}\big)^2}{\beta_i^{(d)}}
      D_\mathrm{PE}\left(
        \mu_{\gamma^\prime,\,i}^{(d)} \,\middle\|\, \mu_{\gamma,\,i}^{(d)}
      \right)
    }_{\text{within-regime divergence}}
    +
    \underbrace{
      D_\mathrm{PE}\left(\alpha^{(d)} \middle\|\, \beta^{(d)}\right)
    }_{\text{inter-regime divergence}}
    \hspace{-12pt},\hspace{-2pt}
  \end{aligned}
  \label{eq:ped-regime-decomp}
\end{equation}
where $\alpha^{(d)} \coloneqq \big(\alpha_i^{(d)}\big)_{i=1}^{K^{(d)}}$ and $\beta^{(d)} \coloneqq \big(\beta_i^{(d)}\big)_{i=1}^{K^{(d)}}$ are the discrete distributions over regimes.
\end{lemma}

The proof is provided in \cref{apx:proof-theorem-ped-regime}.
The absolute continuity $\mu_{\gamma^\prime}^{(d)} \ll \mu_{\gamma}^{(d)}$ holds for the extended marginals in the same way as in \cref{eq:ped-local-hpi}.
The regime-wise $\smash{\mu_{\gamma^\prime,\,i}^{(d)} \ll \mu_{\gamma,\,i}^{(d)}}$ also holds for all regimes with $\beta_i^{(d)}>0$,
since absolute continuity is preserved under restriction to measurable subsets.
If a regime $i$ corresponds to an inactive configuration
(\ie, $\mathcal{Z}_i^{(d)}=\{\bot\}$), then $\mu_{\gamma^\prime,\,i}^{(d)}$ and $\mu_{\gamma,\,i}^{(d)}$ are both degenerate at $\bot$, so $D_{\mathrm{PE}}\left(\mu_{\gamma^\prime,\,i}^{(d)}\,\middle\|\,\mu_{\gamma,\,i}^{(d)}\right)=0$ and only the inter-regime term contributes.

Using this decomposition, we obtain \cref{eq:standard-local-marginal-variance}.

\subsection{Proof of \cref{thm:ped-regime}}
\label{apx:proof-theorem-ped-regime}

\begin{proof}
Here, we omit the superscript ${(d)}$ for notational simplicity.
Since $\mu_{\gamma^\prime} \ll \mu_{\gamma}$ and $\mu_{\gamma^\prime,\,i}\ll\mu_{\gamma,\,i}$ holds for $\beta_i>0$, the
Radon--Nikod\'ym derivative $d\mu_{\gamma^\prime}/d\mu_{\gamma}$ exists and is given by:
\begin{equation}
  \frac{d\mu_{\gamma^\prime}}{d\mu_{\gamma}}(x)
  =
  \frac{d(\alpha_i\,\mu_{\gamma^\prime,\,i})}{d(\beta_i\,\mu_{\gamma,\,i})}(x)
  =
  \frac{\alpha_i}{\beta_i}\,\frac{d\mu_{\gamma^\prime,\,i}}{d\mu_{\gamma,\,i}}(x),
\end{equation}
for $x\in A_i$.
We denote $s_i \coloneqq d\mu_{\gamma^\prime,\,i}/d\mu_{\gamma,\,i}$.

The Pearson divergence can be expressed as follows:
\begin{equation}
    \begin{aligned}
      D_{\mathrm{PE}}(\mu_{\gamma^\prime}\|\mu_{\gamma})
     & =
      \int\left(
        \frac{
          d\mu_{\gamma^\prime}
        }{
          d\mu_{\gamma}
        }
        -1
      \right)^{2}
      d\mu_{\gamma}
      \\
      &=
      \sum_{i=1}^{\;K^{(d)}}
      \int_{A_i^{(d)}}\!\Bigl(\frac{\alpha_i}{\beta_i} s_i - 1\Bigr)^{2} d(\beta_i\,\mu_{\gamma,\,i})
      \\
      &=
      \sum_{i=1}^{\;K^{(d)}}
      \beta_i
      \int\!\Bigl(\frac{\alpha_i}{\beta_i} s_i - 1\Bigr)^{2} d\mu_{\gamma,\,i}
      \\
      &=
      \sum_{i=1}^{\;K^{(d)}}
      \left(
        \frac{\alpha_i^2}{\beta_i} \int s_i^{2} d\mu_{\gamma,\,i}
        - 2\alpha_i \int s_i d\mu_{\gamma,\,i}
        + \beta_i
      \right).
      \label{eq:proof-ped-regime-0}
    \end{aligned}
\end{equation}
Here, we have:
\begin{align}
  \int s_i d\mu_{\gamma,\,i}
  = \int \frac{d\mu_{\gamma^\prime,\,i}}{d\mu_{\gamma,\,i}} d\mu_{\gamma,\,i}
  = \mu_{\gamma^\prime,\,i}(A_i) = 1.
  \label{eq:proof-ped-regime-1}
\end{align}
Thus, we obtain:
\begin{equation}
    \begin{aligned}
      \int s_i^2 d\mu_{\gamma,\,i}
      &=
      \int s_i^2 d\mu_{\gamma,\,i}
      - 2 \int s_i d\mu_{\gamma,\,i} + 1 + 1 \\
      &=
      \int (s_i - 1)^2 d\mu_{\gamma,\,i} + 1 \\
      &=
      D_{\mathrm{PE}}\left(\mu_{\gamma^\prime,\,i}\,\middle\|\,\mu_{\gamma,\,i}\right) + 1.
      \label{eq:proof-ped-regime-2}
    \end{aligned}
\end{equation}
By substituting \cref{eq:proof-ped-regime-1,eq:proof-ped-regime-2}
into \cref{eq:proof-ped-regime-0}, we have:
\begin{equation}
  \begin{aligned}
    D_{\mathrm{PE}}(\mu_{\gamma^\prime}\|\mu_{\gamma})
    & =
    \sum_{i=1}^{\;K^{(d)}}
    \left(
      \frac{\alpha_i^2}{\beta_i}
      \left(
        D_{\mathrm{PE}}\left(\mu_{\gamma^\prime,\,i}\,\middle\|\,\mu_{\gamma,\,i}\right) + 1
      \right)
      - 2\alpha_i + \beta_i
    \right)\\
    & =
    \sum_{i=1}^{\;K^{(d)}}
    \left(
      \frac{\alpha_i^2}{\beta_i}
      D_{\mathrm{PE}}\left(
        \mu_{\gamma^\prime,\,i}\,\middle\|\,\mu_{\gamma,\,i}
      \right)
      +
      \frac{(\alpha_i - \beta_i)^2}{\beta_i}
    \right)\\
    & =
    \sum_{i=1}^{\;K^{(d)}}
    \frac{\alpha_i^2}{\beta_i}
    D_{\mathrm{PE}}\left(
      \mu_{\gamma^\prime,\,i}\,\middle\|\,\mu_{\gamma,\,i}
    \right)
    +
    \sum_{i=1}^{\;K^{(d)}}
    \frac{(\alpha_i - \beta_i)^2}{\beta_i}
    \\
    &=
    \sum_{i=1}^{\;K^{(d)}} \frac{\alpha_i^2}{\beta_i}
    D_{\mathrm{PE}}\left(
      \mu_{\gamma^\prime,\,i} \,\middle\|\, \mu_{\gamma,\,i}
    \right)
    +
    D_{\mathrm{PE}}\left(
      \alpha \,\middle\|\, \beta
    \right).
  \end{aligned}
\end{equation}
This completes the proof.
\end{proof}

\section{Additional Results and Discussion on Experiments on Real-World Benchmarks (\cref{sec:experiments-real-world})}
\label{apx:additional-results-real-world}

This section provides additional real-world results that complement the main experiments in \cref{sec:experiments-real-world}.

\subsection{\methodname Results on Other Instances}
\label{apx:additional-results-real-world-other-instances}

\begin{figure*}[t]
  \centering
  \captionsetup[sub]{skip=-5pt}
  \begin{minipage}{\textwidth}
    \hspace{-3mm}\includegraphics[width=\textwidth]{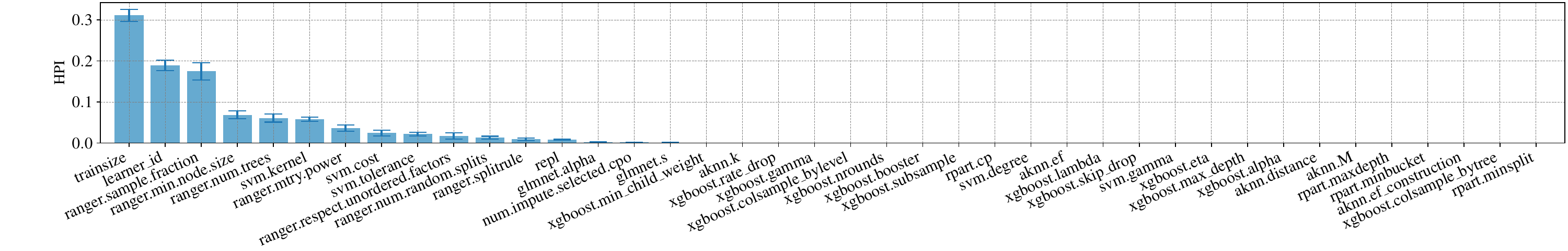}
    \subcaption{Instance ID 1457}
  \end{minipage}
  \par\vspace{2mm}
  \begin{minipage}{\textwidth}
    \hspace{-3mm}\includegraphics[width=\textwidth]{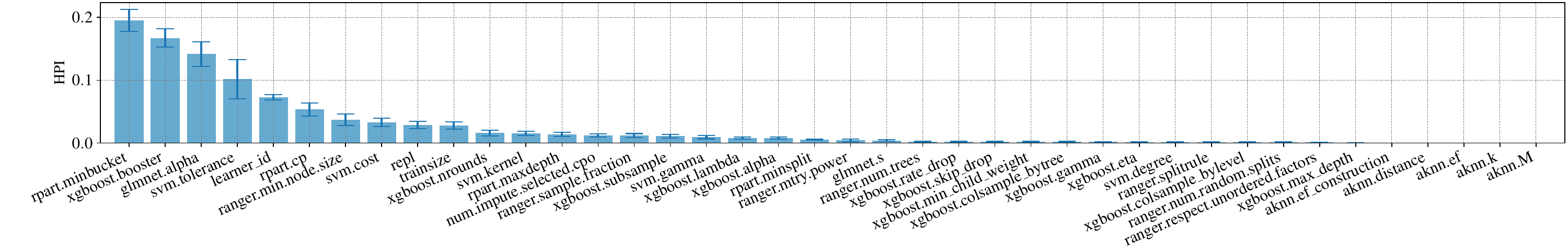}
    \subcaption{Instance ID 1063}
  \end{minipage}
  \par\vspace{2mm}
  \begin{minipage}{\textwidth}
    \hspace{-3mm}\includegraphics[width=\textwidth]{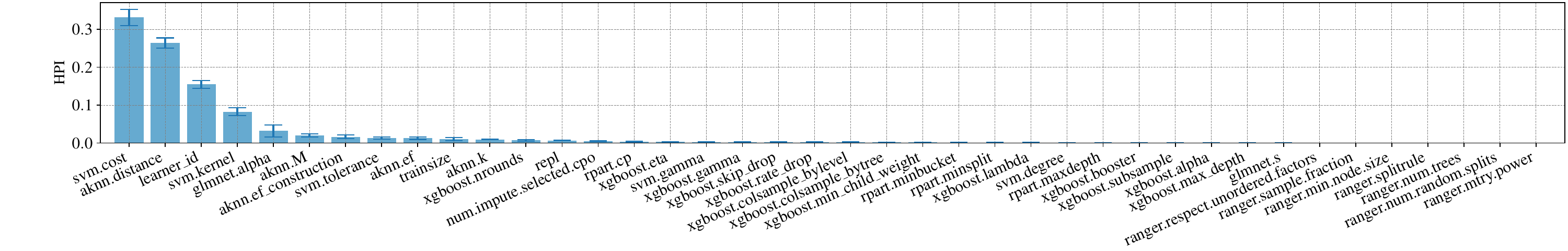}
    \subcaption{Instance ID 1479}
  \end{minipage}
  \par\vspace{2mm}
  \begin{minipage}{\textwidth}
    \hspace{-3mm}\includegraphics[width=\textwidth]{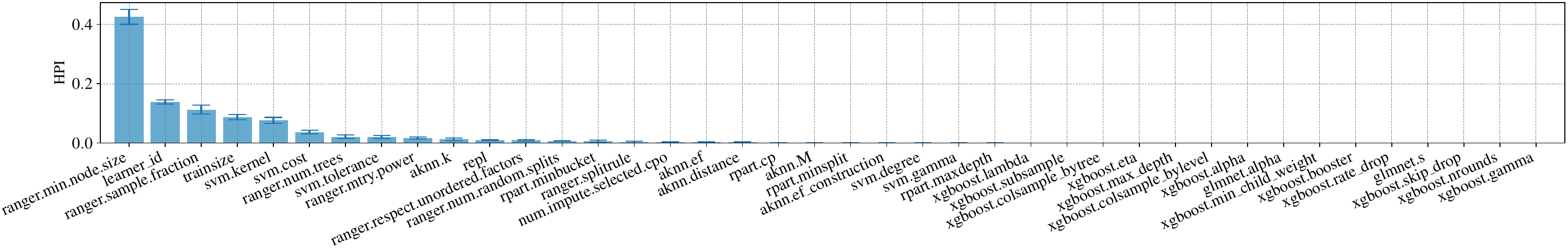}
    \subcaption{Instance ID 15}
  \end{minipage}
  \par\vspace{2mm}
  \begin{minipage}{\textwidth}
    \hspace{-3mm}\includegraphics[width=\textwidth]{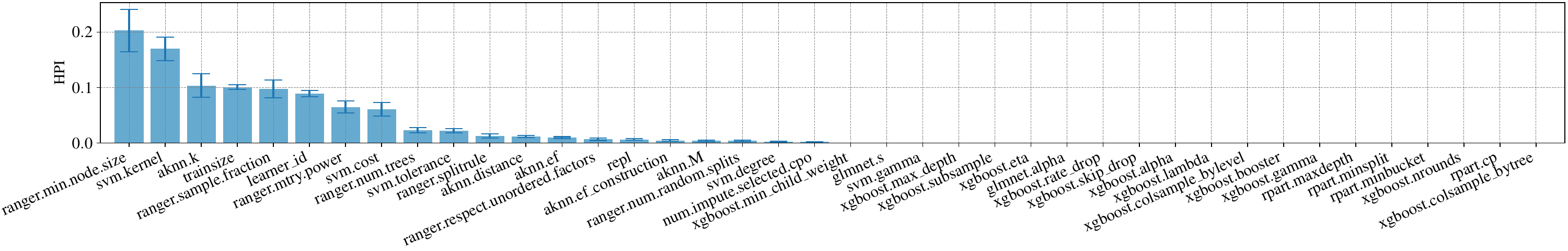}
    \subcaption{Instance ID 1468}
  \end{minipage}

  \vspace{-3mm}
  \caption{
    \methodname HPIs on real-world problems from the YAHPO Gym \texttt{rbv2\_super} scenario with other instances~\citep{pmlr-v188-pfisterer22a,binder2020collecting}.
    The bars denote the mean, and the error bars denote the standard error, both computed over ten independent runs with different random seeds.
  }
  \label{fig:yahpo-rbv2-super-other-instances-cond-ped-anova}
\end{figure*}

In \cref{sec:experiments-real-world}, we presented results on the YAHPO Gym \texttt{rbv2\_super} scenario (instance ID 1053). Here, we report additional results on other instances of the \texttt{rbv2\_super} benchmark suite provided by YAHPO Gym.
The results are shown in \cref{fig:yahpo-rbv2-super-other-instances-cond-ped-anova}.
The corresponding statistics of objective values (\texttt{acc}) for each \texttt{learner\_id} are provided in \cref{tbl:rbv2-super-acc-by-learner-other-instances}.

These results show that, across instances, \texttt{learner\_id}, \texttt{train\allowbreak\_size}, and hyperparameters associated with the top-performing learners occupy the top ranks in HPI.
Moreover, the HPIs produced by \methodname broadly match the learner-wise performance ordering in \cref{tbl:rbv2-super-acc-by-learner-other-instances}, supporting their validity.
In instance ID 1468, although \texttt{aknn} attains the highest mean performance, \texttt{svm} and \texttt{ranger} achieve higher maximum performance. This is reflected in higher HPIs in \texttt{svm} and \texttt{ranger}, which is reasonable given that our goal is to achieve the highest possible score.

In instance IDs 1063 and 1468, the performance gaps among the top learners (in terms of mean and maximum accuracy) are smaller than in instance 1457. Consistently, for these instances, the HPI of \texttt{learner\_id} is slightly lower than that of hyperparameters associated with the top learners, which is reasonable.

Moreover, across all instances, \methodname consistently assigns high HPI to empirically known important hyperparameters within each learner, such as \texttt{sample.fraction} and \texttt{min.node.size} for \texttt{ranger}, the \texttt{cost} and \texttt{kernel} for \texttt{svm}, and \texttt{k} for \texttt{aknn}.
This is reasonable given their well-known influence on model performance.

These results demonstrate that \methodname consistently yields meaningful and interpretable HPIs across multiple real-world instances of the \texttt{rbv2\_super} suite, emphasizing its robustness and applicability in practical scenarios.

\begin{table}[t!]
  \centering
  \caption{
    Summary statistics of the objective value (\texttt{acc}) for each \texttt{learner\_id} in other instances~\citep{pmlr-v188-pfisterer22a,binder2020collecting}.
    Values are reported as mean $\pm$ standard error, both computed over 10 independent runs with different random seeds. The \texttt{learner\_id}s are sorted from best to worst max performance.
}
\vspace{-3mm}
  \label{tbl:rbv2-super-acc-by-learner-other-instances}
  \par\vspace{2mm}
  \begin{minipage}{\linewidth}
    \centering
    \subcaption{Instance ID 1457}
      \begin{tabular}{lccc}
      \toprule
      \textbf{Learner ID} & \textbf{Min} & \textbf{Mean} & \textbf{Max} \\
      \midrule
      \texttt{ranger}  & $0.014 \pm 0.001$ & $0.555 \pm 0.004$ & $0.832 \pm 0.003$ \\
      \texttt{svm}     & $0.000 \pm 0.000$ & $0.211 \pm 0.008$ & $0.803 \pm 0.003$ \\
      \texttt{glmnet}  & $0.007 \pm 0.000$ & $0.146 \pm 0.002$ & $0.657 \pm 0.019$ \\
      \texttt{rpart}   & $0.003 \pm 0.000$ & $0.063 \pm 0.002$ & $0.362 \pm 0.006$ \\
      \texttt{aknn}    & $0.014 \pm 0.000$ & $0.132 \pm 0.002$ & $0.354 \pm 0.003$ \\
      \texttt{xgboost} & $0.008 \pm 0.001$ & $0.046 \pm 0.001$ & $0.354 \pm 0.017$ \\
      \bottomrule
    \end{tabular}
  \end{minipage}
  \par\vspace{2mm}
  \begin{minipage}{\linewidth}
    \centering
    \subcaption{Instance ID 1063}
      \begin{tabular}{lccc}
        \toprule
        \textbf{Learner ID} & \textbf{Min} & \textbf{Mean} & \textbf{Max} \\
        \midrule
        \texttt{svm}     & $0.711 \pm 0.006$ & $0.816 \pm 0.001$ & $0.933 \pm 0.005$ \\
        \texttt{xgboost} & $0.524 \pm 0.008$ & $0.746 \pm 0.001$ & $0.902 \pm 0.002$ \\
        \texttt{rpart}   & $0.733 \pm 0.002$ & $0.818 \pm 0.001$ & $0.897 \pm 0.003$ \\
        \texttt{ranger}  & $0.727 \pm 0.001$ & $0.818 \pm 0.000$ & $0.874 \pm 0.002$ \\
        \texttt{glmnet}  & $0.728 \pm 0.003$ & $0.802 \pm 0.001$ & $0.883 \pm 0.002$ \\
        \texttt{aknn}    & $0.187 \pm 0.002$ & $0.649 \pm 0.006$ & $0.846 \pm 0.001$ \\
        \bottomrule
      \end{tabular}
  \end{minipage}
  \par\vspace{2mm}
  \begin{minipage}{\linewidth}
    \centering
    \subcaption{Instance ID 1479}
      \begin{tabular}{lccc}
        \toprule
        \textbf{Learner ID} & \textbf{Min} & \textbf{Mean} & \textbf{Max} \\
        \midrule
        \texttt{aknn}    & $0.449 \pm 0.002$ & $0.630 \pm 0.003$ & $0.987 \pm 0.002$ \\
        \texttt{svm}     & $0.418 \pm 0.003$ & $0.554 \pm 0.003$ & $0.820 \pm 0.009$ \\
        \texttt{xgboost} & $0.406 \pm 0.007$ & $0.524 \pm 0.001$ & $0.652 \pm 0.006$ \\
        \texttt{glmnet}  & $0.430 \pm 0.002$ & $0.513 \pm 0.001$ & $0.646 \pm 0.006$ \\
        \texttt{rpart}   & $0.444 \pm 0.002$ & $0.498 \pm 0.001$ & $0.629 \pm 0.008$ \\
        \texttt{ranger}  & $0.446 \pm 0.005$ & $0.513 \pm 0.001$ & $0.583 \pm 0.005$ \\
        \bottomrule
      \end{tabular}
  \end{minipage}
  \par\vspace{2mm}
  \begin{minipage}{\linewidth}
    \centering
    \subcaption{Instance ID 15}
      \begin{tabular}{lccc}
        \toprule
        \textbf{Learner ID} & \textbf{Min} & \textbf{Mean} & \textbf{Max} \\
        \midrule
        \texttt{ranger}  & $0.588 \pm 0.003$ & $0.905 \pm 0.002$ & $0.985 \pm 0.000$ \\
        \texttt{svm}     & $0.776 \pm 0.008$ & $0.921 \pm 0.001$ & $0.980 \pm 0.001$ \\
        \texttt{aknn}    & $0.312 \pm 0.001$ & $0.721 \pm 0.006$ & $0.972 \pm 0.001$ \\
        \texttt{rpart}   & $0.648 \pm 0.005$ & $0.886 \pm 0.002$ & $0.963 \pm 0.001$ \\
        \texttt{xgboost} & $0.469 \pm 0.007$ & $0.714 \pm 0.002$ & $0.931 \pm 0.003$ \\
        \texttt{glmnet}  & $0.646 \pm 0.004$ & $0.795 \pm 0.001$ & $0.929 \pm 0.001$ \\
        \bottomrule
      \end{tabular}
  \end{minipage}
  \par\vspace{2mm}
  \begin{minipage}{\linewidth}
    \centering
    \subcaption{Instance ID 1468}
      \begin{tabular}{lccc}
        \toprule
        \textbf{Learner ID} & \textbf{Min} & \textbf{Mean} & \textbf{Max} \\
        \midrule
        \texttt{svm}     & $0.049 \pm 0.004$ & $0.471 \pm 0.010$ & $0.968 \pm 0.001$ \\
        \texttt{ranger}  & $0.087 \pm 0.003$ & $0.782 \pm 0.004$ & $0.943 \pm 0.002$ \\
        \texttt{aknn}    & $0.252 \pm 0.014$ & $0.808 \pm 0.002$ & $0.934 \pm 0.002$ \\
        \texttt{glmnet}  & $0.102 \pm 0.002$ & $0.404 \pm 0.005$ & $0.890 \pm 0.003$ \\
        \texttt{rpart}   & $0.083 \pm 0.001$ & $0.309 \pm 0.004$ & $0.786 \pm 0.007$ \\
        \texttt{xgboost} & $0.080 \pm 0.003$ & $0.206 \pm 0.001$ & $0.744 \pm 0.010$ \\
        \bottomrule
      \end{tabular}
  \end{minipage}
\end{table}

\newcommand{\iamlcorrtableheaderfont}{\fontsize{7.5}{8.5}\selectfont}

\begin{table*}[t]
  \centering
  \caption{
    Correlation between learner-level performance and learner-specific HPI estimates on \texttt{iaml\_super}.
    For each instance, we compute the correlation between the maximum objective value for each \texttt{learner\_id} and the highest HPI among the hyperparameters associated with that \texttt{learner\_id}.
  }

  \vspace{-2mm}
  \label{tbl:iaml-correlation-with-learner-performance}
  \vspace{-0.2em}
  \setlength{\tabcolsep}{3.5pt}
  \resizebox{\textwidth}{!}{%
    \begin{tabular}{lccccccccc}
      \toprule
      \textbf{Instance ID} &
      \textbf{\shortstack{\iamlcorrtableheaderfont condPED-ANOVA\\\iamlcorrtableheaderfont (Ours)}} &
      \textbf{\shortstack{\iamlcorrtableheaderfont PED-ANOVA\\\iamlcorrtableheaderfont w/ Filtering}} &
      \textbf{\shortstack{\iamlcorrtableheaderfont PED-ANOVA\\\iamlcorrtableheaderfont w/ Imputation}} &
      \textbf{\shortstack{\iamlcorrtableheaderfont f-ANOVA\\\iamlcorrtableheaderfont w/ Filtering}} &
      \textbf{\shortstack{\iamlcorrtableheaderfont f-ANOVA\\\iamlcorrtableheaderfont w/ Imputation}} &
      \textbf{\shortstack{\iamlcorrtableheaderfont MDI\\\iamlcorrtableheaderfont w/ Filtering}} &
      \textbf{\shortstack{\iamlcorrtableheaderfont MDI\\\iamlcorrtableheaderfont w/ Imputation}} &
      \textbf{\shortstack{\iamlcorrtableheaderfont SHAP\\\iamlcorrtableheaderfont w/ Filtering}} &
      \textbf{\shortstack{\iamlcorrtableheaderfont SHAP\\\iamlcorrtableheaderfont w/ Imputation}} \\
      \midrule
      40981 & $\bm{0.83}$ & $-0.06$ & $0.78$ & $-0.78$ & $-0.17$ & $0.08$ & $-0.05$ & $0.18$ & $0.17$ \\
      41146 & $\bm{0.42}$ & $-0.98$ & $0.35$ & $0.07$ & $0.09$ & $-0.97$ & $-0.02$ & $-0.60$ & $-0.31$ \\
      1489 & $\bm{0.85}$ & $0.34$ & $0.43$ & $0.22$ & $0.02$ & $0.28$ & $0.27$ & $0.52$ & $0.60$ \\
      1067 & $\bm{0.78}$ & $-0.54$ & $-0.05$ & $0.52$ & $-0.06$ & $0.75$ & $0.04$ & $0.16$ & $0.12$ \\
      \midrule
      Mean $\pm$ StdErr &
      $\bm{0.72 \pm 0.09}$ &
      $-0.31 \pm 0.25$ &
      $0.37 \pm 0.15$ &
      $0.01 \pm 0.24$ &
      $-0.03 \pm 0.05$ &
      $0.03 \pm 0.31$ &
      $0.06 \pm 0.06$ &
      $0.07 \pm 0.20$ &
      $0.15 \pm 0.16$ \\
      \bottomrule
    \end{tabular}%
  }
\end{table*}

\subsection{Comparison with Existing Methods on Another Scenario}
\label{apx:additional-results-real-world-comparison-other-scenario}

In \cref{sec:experiments-real-world-comparison}, we compared \methodname with naive extensions of existing HPI estimators on the \texttt{rbv2\_super} scenario. Here, we report additional results on the \texttt{iaml\_super} scenario in YAHPO Gym.
Similar to the \texttt{rbv2\_super} scenario, the \texttt{iaml\_super} scenario is also a model selection problem, where the conditional hyperparameter \texttt{learner\_id} selects among \texttt{glmnet}, \texttt{ranger}, \texttt{rpart}, and \texttt{xgboost}.

The results are shown in \cref{tbl:iaml-correlation-with-learner-performance}.
\methodname achieves the highest correlation on all instances, indicating that it reliably captures the conditional structure induced by \texttt{learner\_id}.
In contrast, several baselines exhibit unstable behavior, including negative correlations on some instances, whereas \methodname maintains consistently positive and high correlations across the \texttt{iaml} instances.

\end{document}